\documentclass[10pt,twocolumn,letterpaper]{article}

\usepackage[pagenumbers]{cvpr} %

\usepackage[accsupp]{axessibility}
\usepackage{times}
\usepackage{epsfig}
\usepackage{graphicx}
\usepackage{amsmath}
\usepackage{amssymb}

\usepackage{mathtools}  %
\usepackage{url}            %
\usepackage{booktabs}       %
\usepackage{amsfonts}       %
\usepackage{nicefrac}       %
\usepackage{microtype}      %
\usepackage{wrapfig}
\usepackage{subcaption} 
\usepackage{siunitx}

\usepackage[pagebackref=true,breaklinks=true,letterpaper=true,colorlinks,bookmarks=false]{hyperref}

\usepackage[capitalise]{cleveref}
\crefname{section}{Sec.}{Secs.}
\Crefname{section}{Sec.}{Sections}
\crefname{subsection}{Sec.}{Secs.}
\Crefname{subsection}{Section}{Sections}
\Crefname{table}{Table}{Tables}
\crefname{table}{Tab.}{Tabs.}
\creflabelformat{equation}{#2#1#3}  %
\crefname{equation}{Eq.}{Eq.}

\def\cvprPaperID{5244} %
\def\confName{CVPR}
\def\confYear{2022}

\newcommand{\methodname}{CALVIN}

\newcommand{\Afinished}{D}  %

\newcommand{\Ssuccess}{W}  %
\newcommand{\Sincorrect}{F} %
\newcommand{\Rincorrect}{R_F}  %

\newcommand{\tpm}[1]{\tiny$\pm$#1}

\newcommand{\Apred}{\widehat{A}}
\newcommand{\Ppred}{\widehat{P}}
\newcommand{\Rpred}{\widehat{R}}

\newcommand{\Apredfunc}{\Apred(s,a)}
\newcommand{\Ppredfunc}{\Ppred(s'-s|a)}
\newcommand{\Rpredfunc}{\Rpred(a, s'-s)}
\newcommand{\Rpredincorrect}{\widehat{R}_F}

\newcommand{\Apredvalid}{\Apred_\text{logit}}  %

\renewcommand{\paragraph}[1]{\vspace{2pt plus 1pt minus 1pt}\noindent{\bf #1}\;}

\makeatletter
\def\@seccntformat#1{\@ifundefined{#1@cntformat}%
  {\csname the#1\endcsname\quad}  %
  {\csname #1@cntformat\endcsname}%
}
\let\oldappendix\appendix %
\renewcommand\appendix{%
    \oldappendix
    \newcommand{\section@cntformat}{\appendixname~\thesection\quad}
}
\makeatother

\begin{document}

\title{Towards real-world navigation with deep differentiable planners}

\author{Shu Ishida \hspace{0.5in} Jo\~{a}o F. Henriques\\
Visual Geometry Group\\
University of Oxford\\
{\tt\small \{ishida, joao\}@robots.ox.ac.uk}
}

\maketitle

\begin{abstract}

We train embodied neural networks to plan and navigate unseen complex 3D environments, emphasising real-world deployment.
Rather than requiring prior knowledge of the agent or environment, the planner learns to model the state transitions and rewards.
To avoid the potentially hazardous trial-and-error of reinforcement learning, we focus on differentiable planners such as Value Iteration Networks (VIN), which are trained offline from safe expert demonstrations. 
Although they work well in small simulations, we address two major limitations that hinder their deployment.
First, we observed that current differentiable planners struggle to plan long-term in environments with a high branching complexity.
While they should ideally learn to assign low rewards to obstacles to avoid collisions,
these penalties are not strong enough to guarantee collision-free operation.
We thus impose a structural constraint on the value iteration, which explicitly learns to model impossible actions and noisy motion.
Secondly, we extend the model to plan exploration with a limited perspective camera under translation and fine rotations, which is crucial for real robot deployment.
Our proposals significantly improve semantic navigation and exploration on several 2D and 3D environments, succeeding in settings that are otherwise challenging for differentiable planners.
As far as we know, we are the first to successfully apply them to the difficult Active Vision Dataset, consisting of real images captured from a robot.\footnote{Code available: \url{https://github.com/shuishida/calvin}}

\end{abstract}

\begin{figure}
  \centering
  \renewcommand*{\arraystretch}{0}
  \begin{tabular}{*{3}{@{}c}@{}}
  \includegraphics[height=2.15cm]{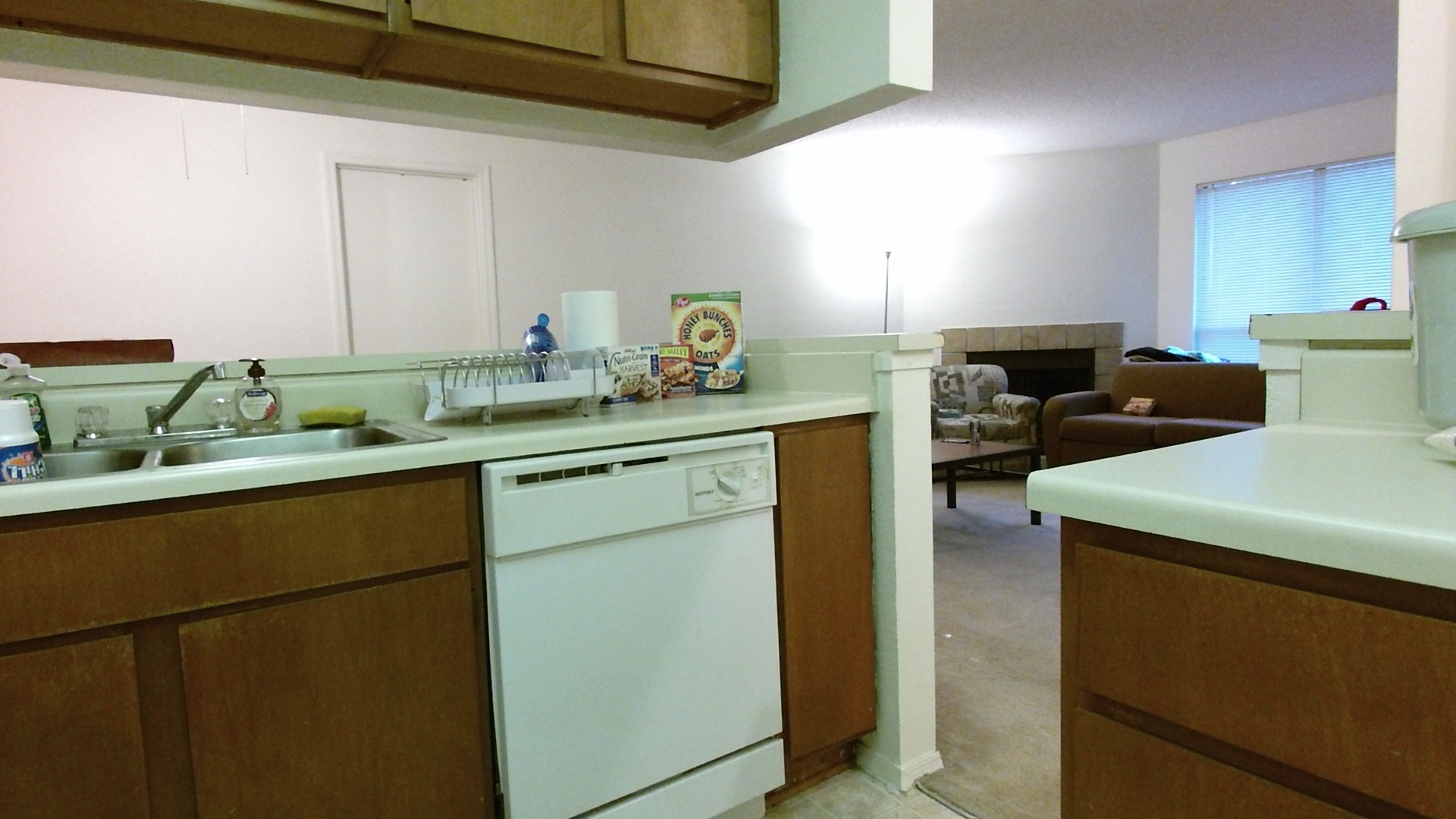} &
  \includegraphics[height=2.15cm]{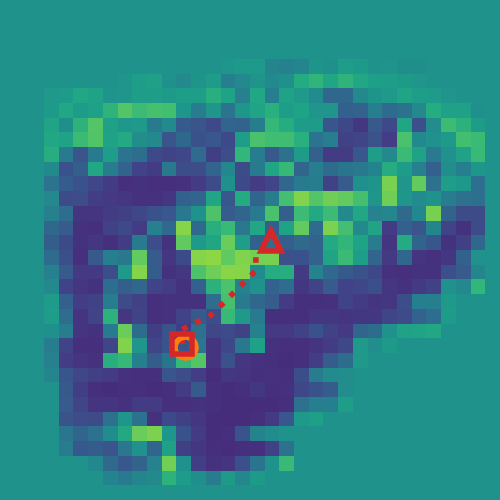} &
  \includegraphics[height=2.15cm]{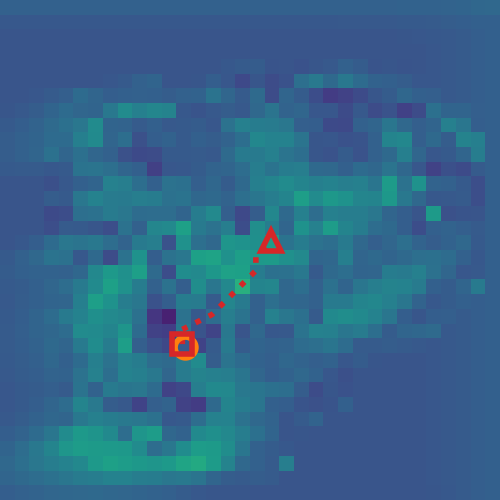} \\
  \includegraphics[height=2.15cm]{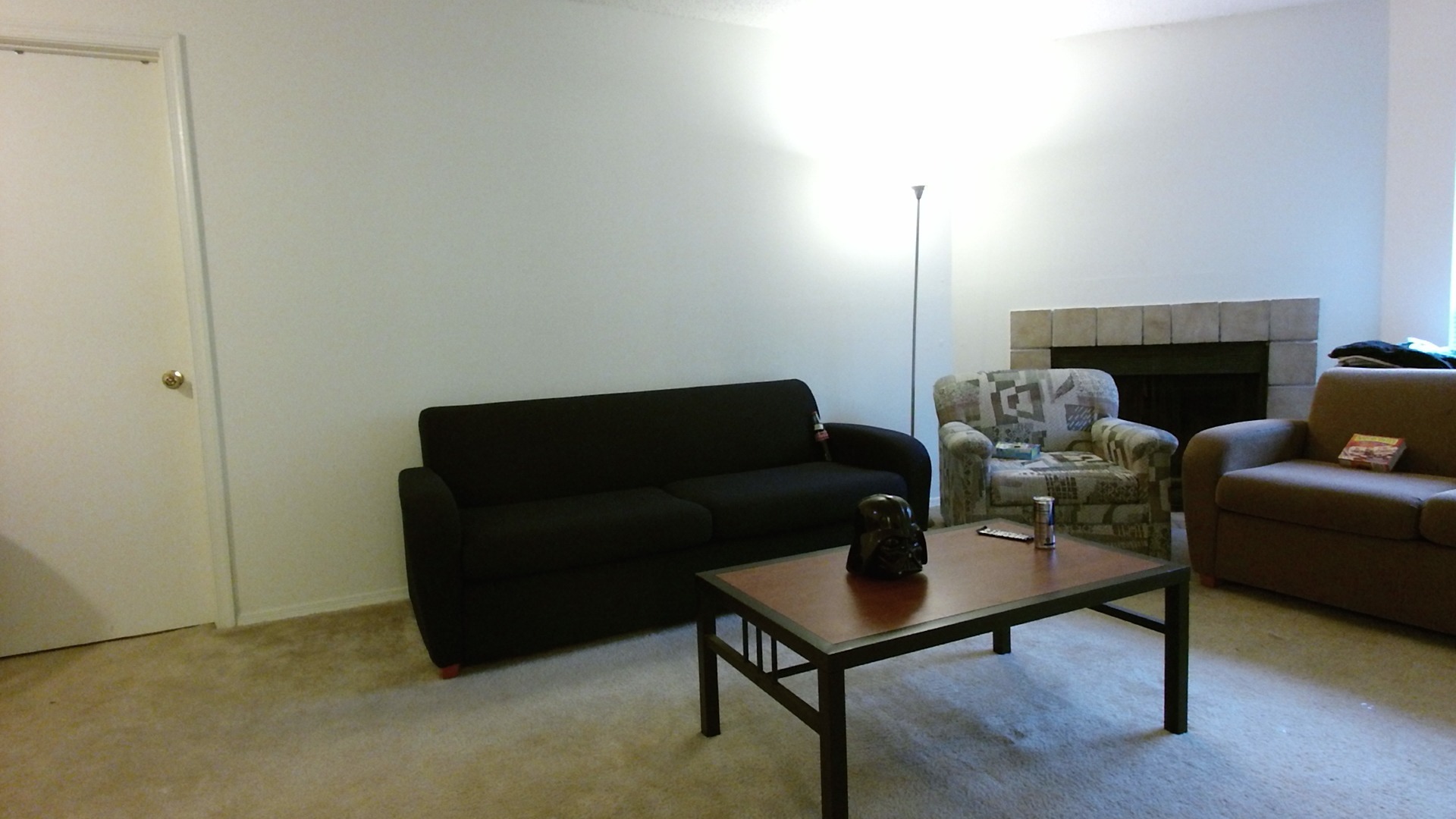} &
  \includegraphics[height=2.15cm]{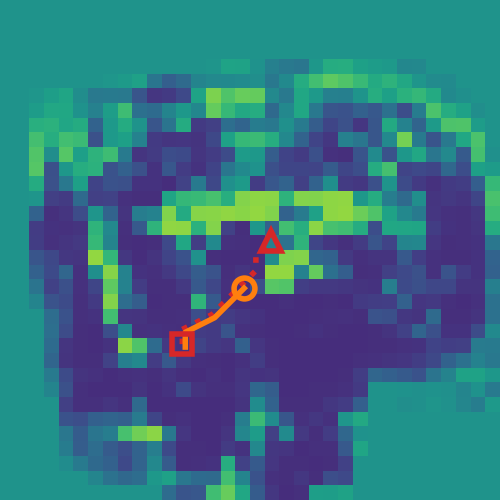} &
  \includegraphics[height=2.15cm]{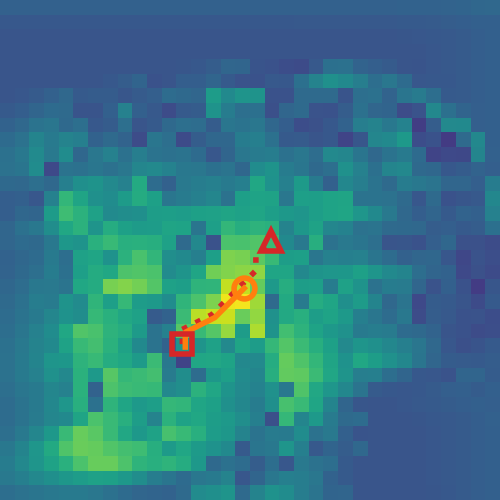} \\
  \includegraphics[height=2.15cm]{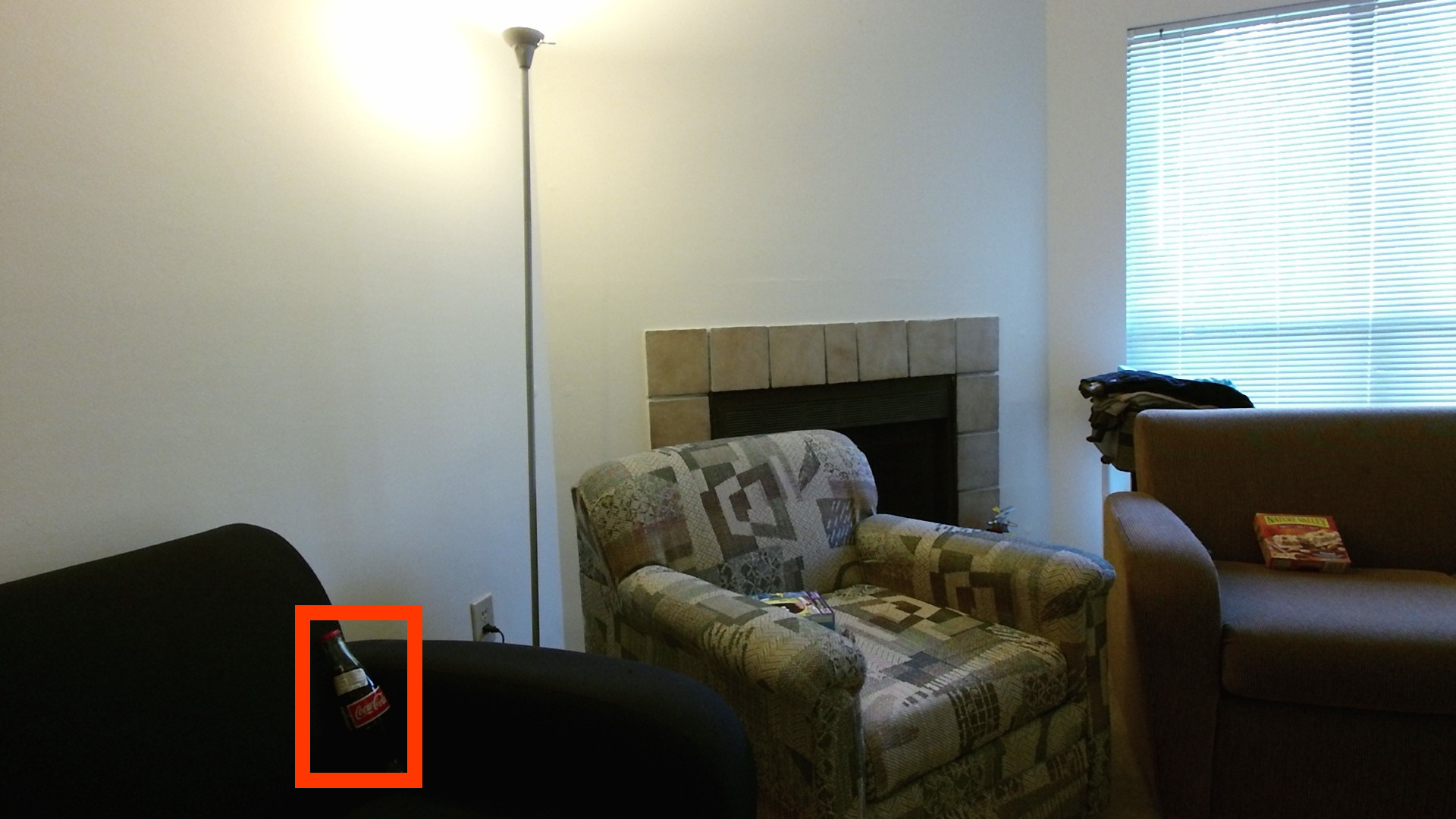} &
  \includegraphics[height=2.15cm]{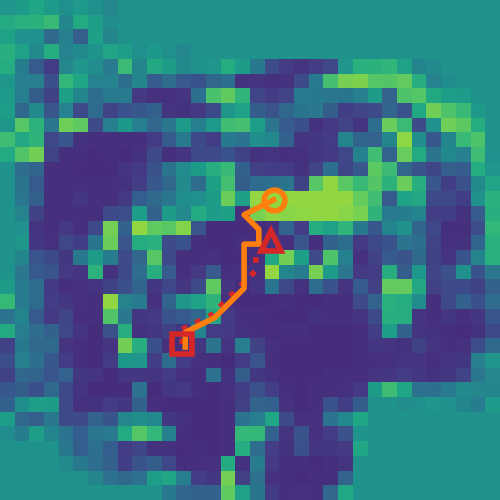} &
  \includegraphics[height=2.15cm]{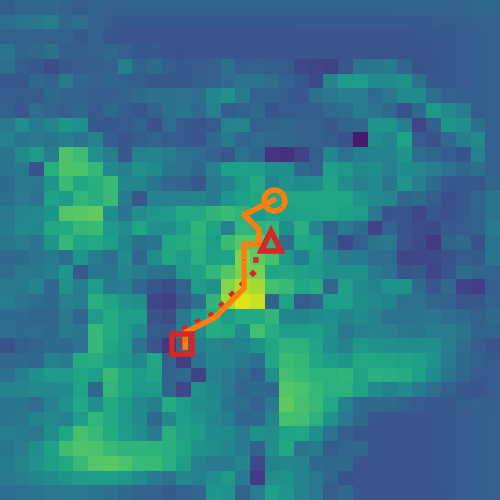}
  \end{tabular}
  \caption{
    \textbf{(1\textsuperscript{st} column)} Input images seen during a run of our method on AVD (\cref{sec:avd}). This embodied neural network has learned to efficiently explore and navigate unseen indoor environments, to seek objects of a given class (highlighted in the last image). \textbf{(2\textsuperscript{nd}-3\textsuperscript{rd} columns)} Predicted rewards and values (resp.), for each spatial location (higher for brighter values). The unknown optimal trajectory is dashed, while the robot's trajectory is solid.
  }
  \label{fig:avd}
\end{figure}

\section{Introduction}
\label{introduction}
Continuous advances in robotics have enabled robots to be deployed to a wide range of scenarios, from manufacturing in factories and cleaning in households, to the emerging applications of autonomous vehicles and delivery drones~\cite{anderson2018evaluation}.
Improving their autonomy is met with many challenges, due to the difficulty of planning from uncertain sensory data.
In classical robotics, the study of planning has a long tradition~\cite{thrun2002probabilistic}, using detailed knowledge of a robot's configuration and sensors, with little emphasis on learning from data.
An almost orthogonal approach is to use deep learning, an intensely data-driven, non-parametric approach~\cite{goodfellow2016deep}.
Modern deep neural networks excel at pattern recognition~\cite{sharif2014cnn}, although they do not offer a direct path to planning applications.
While one approach would be to parse a scene into pre-defined elements (\eg object classes and their poses) to be passed to a more classical planner, an end-to-end approach where all modules are learnable has the chance to improve with data, and be adaptable to novel settings with no manual tuning.
Because of the data-driven setup, a deep network has the potential to learn behaviour that leverages the biases of the environment, such as likely locations for certain types of rooms.
Value Iteration Networks (VINs)~\cite{tamar_value_2016} emerged as an elegant way to merge classical planning and data-driven deep networks, by defining a \emph{differentiable} planner.
Being sub-differentiable, like all other elements of a deep network, allows the planner to include learnable elements, trained end-to-end from example data.
For example, it can learn to identify and avoid obstacles, and to recognise and seek classes of target objects, without explicitly labelled examples.
However, there are gaps between VIN’s idealised formulation and realistic robotics scenarios, which some works address~\cite{gupta_cognitive_2017, karkus_qmdp-net_2017}.
The CNN-based VIN~\cite{tamar_value_2016} considers that the full environment is visible and expressible as a 2D grid. As such, it does not account for embodied (first-person) observations in 3D spaces, unexplored and partially-visible environments, or the mismatch between egocentric observations and idealised world-space discrete states.

In this paper we address these challenges, and close the gap between current differentiable planners and realistic robot navigation applications.
Our contributions are:

\smallskip\noindent
1. A constrained transition model for value iteration, following a rigorous probabilistic formulation, which explicitly models illegal actions and task termination (\cref{sec:augmented}).
This is our main contribution.

\smallskip\noindent
2. A 3D state-space for embodied planning through the robot's translation and rotation (\cref{sec:orientations}).
Planning through \emph{fine-grained} rotations is often overlooked (\cref{sec:related_work}), requiring better priors for transition modelling (\cref{sec:transition-model}).

\smallskip\noindent
3. A trajectory reweighting scheme that addresses the unbalanced nature of navigation training distributions (\cref{sec:reweighting}).

\smallskip\noindent
4. 
We demonstrate for the first time that differentiable planners can learn to navigate in both complex 3D mazes, and the challenging Active Vision Dataset~\cite{ammirato2017dataset}, with images from a real robot platform (\cref{sec:avd}).
Thus our method can be trained with limited data collected offline, as opposed to limitless data from a simulator as in prior work.

\smallskip
\Cref{sec:related_work} discusses related work, while \cref{sec:value_iteration} gives a short introduction to differentiable planning.
\Cref{sec:method} presents our technical proposals, and \cref{sec:experiments} evaluates them.

\section{Related work}
\label{sec:related_work}

\paragraph{Planning and reinforcement learning.}
Planning in fully-known deterministic state spaces was partially solved by graph-search algorithms \cite{dijkstra,a_star,stentz_optimal_1994,prm,rrt1,rrt2}.
A Markov Decision Process (MDP) \cite{bellman_markovian_1957,sutton_reinforcement_2018}
considers probabilistic transitions and rewards, enabling planning on stochastic models and noisy environments.
With known MDPs, Value Iteration achieves the optimal solution~\cite{bellman_markovian_1957,sutton_reinforcement_2018}.
Reinforcement Learning (RL) focuses on unknown MDPs
\cite{watkins_learning_1989,rummery_online_1994,williams_simple_1992,mnih_asynchronous_2016,mnih_asynchronous_2016}.
Model-free RL systems are reactive; a policy (\ie a generic plan) is built over a long period of trial and error, and will be specific to the training environment (there is no online planning).
This makes them less ideal for robotics, where risky failures must be avoided, and new plans (policies) must be created on-the-fly in new environments.
Model-based RL methods attempt to learn an MDP and often use it for planning online~\cite{clavera2018model_mbmpo,janner2019trust_mbpo,kaiser2019model_simple,hafner2019learning_planet,hafner2019dreamer}. The environment is assumed to provide a reward signal at every step in both cases.

\paragraph{Deep networks for navigation.}
Several works have made advances into training deep networks for navigation.
The Neural Map~\cite{parisotto2017neuralmap} is an A2C~\cite{mnih_asynchronous_2016} agent (thus reactive) that reads and writes to a differentiable memory (\ie a map).
Similarly, Mirowski \etal \cite{mirowski_learning_2017} train reactive policies which are specific to an environment.
The Value Prediction Network~\cite{vpn2017} learns an MDP and its state-space from data, and plans with a single roll-out over a short horizon.
Hausknecht \etal~\cite{hausknecht2015pomdps} add recurrence to deep Q-learning to address partially-observable environments, but considering only single-frame occlusions.
Several works treat planning as a non-differentiable module, and focus on training neural networks for other aspects of the navigation system.
Savinov \etal \cite{savinov_semi-parametric_2018} do this by composing siamese networks and value estimators trained on proxy tasks, and use an initial map built from footage of a walk through the environment.
Active Neural SLAM~\cite{chaplot_learning_2020} trains a localisation and mapping component (outputting free space and obstacles), a policy network to select a target for the non-differentiable planner, and another to perform low-level control.
Subsequent work~\cite{chaplot2020object} complemented this map with semantic segmentation.
This map is different from ours, which consists of embeddings learned by a differentiable planner, and so are not constrained to correspond to annotated labels.
The later two works navigate successfully in large simulations of 3D-scanned environments, and were transferred to real robots.

\paragraph{Imitation learning.}
To avoid learning by trial-and-error, Imitation Learning instead uses expert trajectories.
Inverse Reinforcement Learning (IRL) \cite{ng_algorithms_2000,ziebart_maximum_2008,ziebart_modeling_2010,ho_generative_2016} achieves this by learning a reward function that best explains the expert trajectories.
However, this generally leads to a difficult nested optimisation and is an ill-posed problem, since many reward functions can explain the same trajectory.

\paragraph{Differentiable planning.}
Tamar \etal \cite{tamar_value_2016} introduced the Value Iteration Network (VIN) for Imitation Learning, which generates a plan online (with value iteration) and back-propagates errors through the plan to train estimators of the rewards (\cref{sec:vin}).
VINs were also applied to localisation from partial observations by Karkus \etal \cite{karkus_qmdp-net_2017}, but assuming a full map of the environment.
Lee \etal~\cite{lee_gated_2018} replaced the maximum over actions in the VI formula with a LSTM. 
The evaluation included an extension to rotation, but assuming a fully-known four directional view for every gridded state, which is often not available in practice.
A subtle difference is that they handle rotation by applying a linear policy layer to the hidden channels of a 2D state-space grid, which is less interpretable than our 3D translation-rotation grid.

Most previous work on deploying VIN to robots \cite{Schleich_2019,nie2021cin} assume an occupancy map and goal map rather than learning the map embedding and goal themselves from data.
To the best of our knowledge, Gupta \etal's Cognitive Mapper and Planner (CMP) \cite{gupta_cognitive_2017} is the only work which evaluated a differentiable planner on real-world data using map embeddings learned end-to-end.
CMP uses a hierarchical VIN~\cite{tamar_value_2016} to plan on larger environments, and updates an egocentric map.
It handles rotation as a warping operation external to the VIN (i.e. it plans in a 2D translation state-space, not 3D translation-rotation space).
Warping an egocentric map per update achieves scalability at the cost of progressive blurring of the embeddings.
CMP was evaluated for rotations of $\ang{90}$ and deterministic motion, rather than noisy translation and rotation.
Our 3D translation-rotation space and learned transition model allows smooth trajectories with finer rotations.
CMP also requires gathering new trajectories online during training (with DAgger~\cite{pmlr-v15-ross11a}), while ours is trained solely with a fixed set of training trajectories,
foregoing the need for a simulator and the associated domain gap.

\section{Background}
\label{sec:value_iteration}
A Markov Decision Process (MDP) \cite{bellman_markovian_1957,sutton_reinforcement_2018} formalises sequential decision-making.
It consists of states $s \in \mathcal{S}$ (\eg locations), actions available at each state $a \in \mathcal{A}(s)$, a reward function $R(s, a, s')$ to be maximised (\eg reaching the target), and a transition probability $P(s'|s, a)$ (the probability of the next state $s'$ given the current state $s$ and action $a$). 
The objective of an agent is to learn a policy $\pi(a|s)$, specifying the probability of choosing action $a$ for any state $s$, chosen to maximise the expected return $G_t$ at every time step $t$. 
A return is a sum of discounted rewards, $G_t = \sum_{k=0}^{\infty}{\gamma^k R_{t+k+1}}$, where a discount factor $\gamma \in (0, 1)$ prevents divergences of the infinite sum.
The value function $V_{\pi}(s) = \mathbb{E}_{\pi}[G_t | s_t = s]$ evaluates future returns from a state, 
while the action-value function $Q_{\pi}(s, a) = \mathbb{E}_{\pi}[G_t | s_t = s, a_t = a]$ considers both a state and the action taken.
An optimal policy $\pi_*$ should then maximise the expected return for all states, \ie $\forall s \in \mathcal{S}, v_*(s) = \max_{\pi}{v_{\pi}(s)}$. 
Value Iteration (VI)~\cite{bellman_markovian_1957,sutton_reinforcement_2018} is an algorithm to obtain an optimal policy, by alternating a refinement of both value ($V$) and action-value ($Q$) function estimates in each iteration $k$.
When $s$ and $a$ are discrete, $Q^{(k)}$ and $V^{(k)}$ can be implemented as simple tables (tensors).
In particular, we will consider as states the cells of a 2D grid, corresponding to discretised locations in an environment, \ie $s=(i,j) \in \mathcal{S}=\{1, \ldots, N\}^2$. Furthermore, transitions are local (only to coordinates offset by $\delta \in K=\{-1, 0, 1\}^2$):
\begin{align}
Q^{(k)}_{a, s} =
\smashoperator{\sum_{\delta\in K}}
{P_{a,\delta,s} \left(R_{a,\delta,s} +
\gamma V_{s+\delta}^{(k-1)}\right)} \notag
\\
\forall a \in \mathcal{A}(s)
,\quad \mathrm{with} \quad
V^{(k)}_{s} = \max_{a \in \mathcal{A}(s)}{Q^{(k)}_{a, s}},
\label{eq:vi-grid}
\end{align}
Note that to avoid repetitive notation $s$ and $\delta$ are 2D indices, so $V$ is a 2D matrix and both $R$ and $P$ are 5D tensors.
The policy $\pi^{(k)}_{s} = \text{argmax}_{a \in \mathcal{A}}{Q^{(k)}_{a, s}}$ simply chooses the action with the highest action-state value.
While the sum in \cref{eq:vi-grid} resembles a convolution, the filters ($P$) are space-varying (depend on $s=(i,j)$), so it is not directly expressible as such.
\Cref{eq:vi-grid} represents the ``ideal'' VI for local motion on a 2D grid, without further assumptions.

\subsection{Value Iteration Network}\label{sec:vin}
While VI guarantees the optimal policy, it requires that the functions for transition probability $P$ and reward $R$ are known (\eg defined by hand).
Tamar \etal \cite{tamar_value_2016} pointed out that all VI operations are (sub-)differentiable, and as such a model of $P$ and $R$ can be \emph{trained from data} by back-propagation.
For the case of planning on a 2D grid (navigation), they related \cref{eq:vi-grid} to a CNN, as:
\begin{align}
&Q^{(k)}_{a, s} =
\smashoperator{\sum_{\delta\in K}}
{\left(P_{a,\delta}^R \widehat{R}_{s+\delta} +
P_{a,\delta}^V V_{s+\delta}^{(k-1))}\right)}, \notag
\\
&\forall a \in \mathcal{A}
,\quad \mathrm{with} \quad
V^{(k)}_{s} = \max_{a \in \mathcal{A}}{Q^{(k)}_{a, s}},
\label{eq:vin}
\end{align}
where $P^R,\,P^V \in \mathbb{R}^{A \times |K|}$ are two learned convolutional filters that represent the transitions ($P$ in \cref{eq:vi-grid}), and $\widehat{R}$ is a predicted 2D reward map.
Note that $\mathcal{A}$ is independent of $s$, \ie all actions are allowed in all states.
This turns out to be detrimental (see next paragraph).
The reward map $\widehat{R}$ is predicted by a CNN, from an input of the same size that represents the available observations. In Tamar \etal's experiments~\cite{tamar_value_2016}, the observations were a fully-visible overhead image of the environment, from which negative rewards such as obstacles and positive rewards such as navigation targets can be located.
Each action channel in $\mathcal{A}$ corresponds to a move in the 2D grid, typically 8-directional or 4-directional.
Eq.~\ref{eq:vin} is attractive, because it can be implemented as a CNN consisting of alternating convolutional layers ($Q$) and max-pooling along the actions (channels) dimension ($V$).

\begin{figure}
  \centering
  \includegraphics[width=0.325\columnwidth]{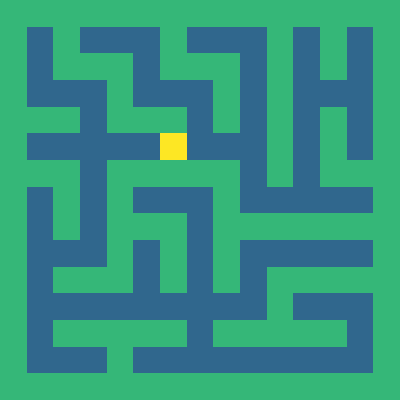}\hfill
  \includegraphics[width=0.325\columnwidth]{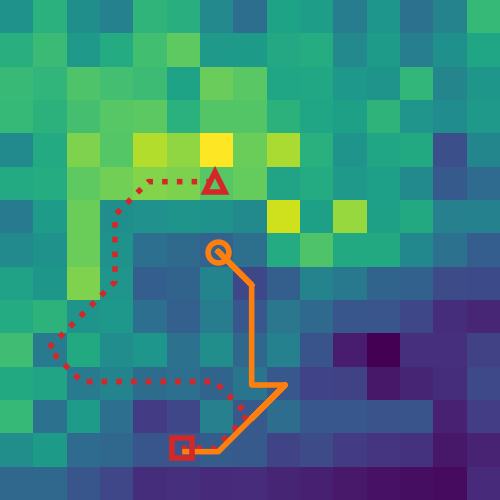}\hfill
  \includegraphics[width=0.325\columnwidth]{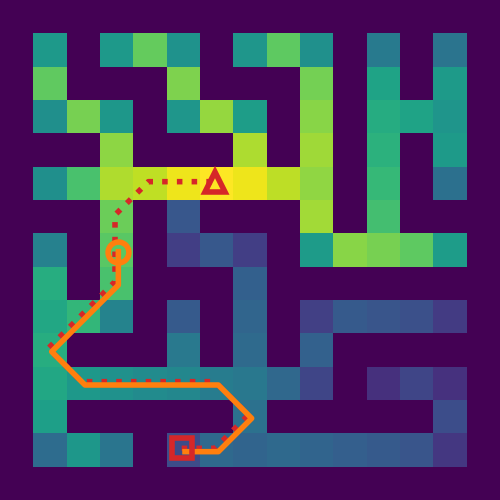}
  \caption{
    \textbf{(left)} A 2D maze, with the target in yellow.
    \textbf{(middle)} Values produced by the VIN for each 2D state (actions are taken towards the highest value). Higher values are brighter. The correct trajectory is dashed, the current one is solid. The agent (orange circle) is stuck due to the local maximum below it.
    \textbf{(right)} Same values for our method, CALVIN. There are no spurious maxima, and the values of walls are correctly considered low (dark).
  }
  \label{fig:values}
\end{figure}

\paragraph{Motivating experiment.}
Since the VIN allows all actions at all states ($\mathcal{A}$ does not depend on $s$), collisions must be modelled as states with low rewards.
In practice, the VIN does not learn to forbid such actions completely, resulting in propagation of values from states which cannot be reached directly due to collision along the way.
We verified this experimentally, by training a VIN according to Tamar \etal~\cite{tamar_value_2016} on 4K mazes (see \cref{sec:experiments} for details).
We then measured, for each state, whether the predicted scores for all valid actions are larger than those for all invalid actions (\ie collisions).
Intuitively, this means the network always prefers free space over collisions.
Surprisingly, we found that this was not true for 24.6\% of the states.
For a real-world robot to work reliably, this is an unacceptably high chance of collisions.
As a comparison, for our method (\cref{sec:method}) this rate is only 1.6\%.
In the same experiment, the VIN often gets trapped in local minima of the value function and does not move (\cref{fig:values}), which is another failure mode.
We aim to fix these issues, and push VINs towards realistic scenarios.
An alternative would be to employ online retraining (DAgger)~\cite{pmlr-v15-ross11a}, which cannot use solely a fixed set of offline trajectories.

\section{Proposed method}
\label{sec:method}

We propose a transition model that accounts for illegal actions and termination.
We then extend it to embodied planning (rigid 3D motion and partially-observed environments).

\subsection{Augmented navigation state-action space}\label{sec:augmented}
In this section we will derive a probabilistic transition model from first principles, with only two assumptions and no extra hyper-parameters.
The first assumption is locality and translation invariance of the agent motion, which was introduced in the VIN to allow efficient learning with shared parameters.
Unlike the VIN, we will decompose the transition model $P(s'|s,a)$ into two components: the agent motion model $\Ppredfunc$, which is translation invariant and shared across states (depending only on the spatial difference between states $s'-s$); and an observation-dependent predictor $\Apredfunc\in[0,1]$ which evaluates whether action $a$ is available from state $s$, to disqualify illegal actions.

In robotics, it is essential that the agent understands that the current task has been completed to move on to the next one.
Since in small environments there is a high chance that a randomly-acting agent will stumble upon the target, Anderson \etal~\cite{anderson2018evaluation} suggested that an explicit termination action must be taken at the target to finish successfully.
Therefore, in addition to positional states, we assume a success state $\Ssuccess$ (``win'') that is reached only by triggering a termination action $\Afinished$ (``done''), and a failure state $\Sincorrect$ (``fail'') that is reached upon triggering an incorrect action.
We denote the reward for reaching $\Sincorrect$ as $\Rincorrect$, and the translation-invariant rewards as $\Rpredfunc$. For simplicity, we consider the reward for success a special case of $\Rpredfunc$ where $a = \Afinished$ and $s' = \Ssuccess$. With these assumptions, the reward function $R(s, a, s')$ is:
\begin{equation}
\label{eq:vinas_r}
R(s, a, s')=
\begin{cases}
    \Rpredincorrect, &s'= \Sincorrect\\
    \Rpredfunc, &s'\neq \Sincorrect.
\end{cases}
\end{equation}
Together with the agent motion model $\Ppredfunc$, action validity predictor $\Apredfunc$ and the definition of a failure state $\Sincorrect$, the transition model $P(s' | s, a)$ can be derived as \cref{eq:vinas_p2}:
\begin{equation}
\label{eq:vinas_p2}
P(s'|s,a)
=\begin{cases}
    1-\Apredfunc, &s'=\Sincorrect\\
    \Apredfunc\Ppredfunc,  &s'\neq \Sincorrect.
\end{cases}
\end{equation}
From the above, we calculate the reward $R(s, a)$ by marginalising over the neighbour states $s'$:
\begin{align} \label{eq:reward_function}
R(s, a) &= \sum_{s'}{P(s'|s,a)R(s,a,s')}\\
= \Rpredincorrect( & 1-\Apredfunc) + \Apredfunc\sum_{s'}{\Ppredfunc\Rpredfunc} \notag
\end{align}
Finally, \cref{eq:reward_function} can be plugged into \cref{eq:vi-grid} to obtain our proposed value iteration's $Q(s,a)$:
\begin{align}
\label{eq:proposed_update}
Q(s, a) &= R(s, a) + \gamma \mathbb{I}_{a\in \mathcal{A}}\sum_{s'}{P(s'|s,a)V(s')}\\
&= R(s, a) + \gamma \Apredfunc\mathbb{I}_{a\neq \Afinished}\sum_{s'}{\Ppredfunc V(s')} \notag
\end{align}
where $\mathbb{I}$ is an indicator function.
\Cref{eq:reward_function,eq:proposed_update} essentially express a constrained VI, which models the case of an MDP on a grid with unknown illegal states and termination at a goal state.
The inputs to this model are three learnable functions ---
the motion model $\Ppredfunc$, the action validity $\Apredfunc$ (\ie obstacle predictions), and the rewards $\Rpredfunc$ and $\Rpredincorrect$.
These are implemented as CNNs with the observations as inputs ($\Rpredincorrect$ is a single learned scalar).
All the constraints follow from a well-defined world-model, with very interpretable predicted quantities, unlike previous proposals~\cite{lee_gated_2018,tamar_value_2016}.
We named this method Collision Avoidance Long-term Value Iteration Network (CALVIN).

\subsubsection{Training}\label{sec:training}
Similarly to Tamar \etal~\cite{tamar_value_2016}, we train our method with a softmax cross-entropy loss $L$, comparing an example trajectory $\{(s_t, a^*_t):t\in T\}$ with predicted action scores $Q(s, a)$:
\begin{equation}\label{eq:training}
    \min_{\Ppred,\,\Apred,\,\Rpred}\tfrac{1}{|T|}
    \textstyle\sum_{t\in T}w_t L\left(Q(s_{t}),\,a_{t}^{*}\right),
\end{equation}
where $Q(s)$ is a vector with one element per action ($Q(s,a)$), and $w_t$ is an optional weight that can be used to bias the loss if $w_t \neq 1$ (\cref{sec:reweighting}).
Example trajectories are shortest paths computed from random starting points to the target (also chosen randomly).
Note that the learned functions can be conditioned on input observations. These are 2D grids of features, with the same size as the considered state-space (\ie a map of observations), which is convenient since it enables $\Ppred$, $\Apred$ and $\Rpred$ to be implemented as CNNs.

\subsubsection{Transition modelling}\label{sec:transition-model}
Similarly to the VIN (\cref{eq:vin}), the motion model $\Ppredfunc$ is implemented as a convolutional filter $\widehat{P} \in \mathbb{R}^{|\mathcal{A}|\times |K|}$, so it only depends on the relative spatial displacement $s'-s$ between the two states $s$ and $s'$.
We can use the transitions already observed in the example trajectory to constrain the model,
by adding a cross-entropy loss term $L(\widehat{P}(a_t^{*}),\,s_{t+1}-s_t)$ for each step in the example trajectory.
After training, the filter $\Ppredfunc$ will consist of a distribution over possible state transitions for each action (visualised in \cref{fig:motion_models}).

\subsubsection{Action availability}\label{sec:action-availability}
Although the available actions $\Apred$ could, in theory, be learned completely end-to-end, in practice we found that additional regularisation is necessary.
If we had a reliable log-probability of each action being taken, $\Apredvalid(s,a)$, then by thresholding it at some point $\Apred_\text{thresh}(s)$ we would distinguish between available and unavailable actions. Using a sigmoid function $\sigma$ as a soft threshold, we can write this as:
\begin{equation}
    \Apred(s, a) = \sigma(\Apredvalid(s,a) - \Apred_\text{thresh}(s)).
\end{equation}
Both $\Apredvalid(s,a)$ and $\Apred_\text{thresh}(s)$ are predicted by the network, given the observations at each time step.
In order to ground the probabilities of each action being taken,
we encourage the actions $\Apredvalid(s_t)$ to match the example trajectory action $a_t^{*}$ for all steps $t$, with an additional cross-entropy loss $L(\Apredvalid(s),\,a^{*})$.
Note that \emph{there is no additional ground truth supervision} -- we use strictly the same data as a VIN.

\subsubsection{Fully vs. partially observable environments}\label{sec:exploration}
Some previous works~\cite{tamar_value_2016,lee_gated_2018} assume that the entire environment is static and fully observable, which is often unrealistic.
Partially observable environments, involving unknown scenes, require exploring to gather information and are thus more challenging.
We account for this with a simple but significant modification.
Note that $Q(s_t)$ in \cref{eq:training} depends on the learned functions (CNNs) $\Ppred$, $\Apred$ and $\Rpred$ through \cref{eq:proposed_update}, and these in turn are computed from the observations.
We extend the VIN framework to the case of partial observations by ensuring that $Q(s_t| O_{\leq t})$ depends \emph{only} on the observations up to time $t$, $ O_{\leq t}$.
This means that the VIN recomputes a plan at each step $t$ (since the observations are different), instead of once for a whole trajectory.
Unobserved locations have their features set to zero, so in practice knowledge of the environment is slowly built up during an expert demonstration, which enables exploration behaviour to be learned.

\subsubsection{Trajectory reweighting}\label{sec:reweighting}
Exploration provides observations $ O_{\leq t}$ of the same locations at different times (\cref{sec:exploration}).
We can thus augment \cref{eq:training} with these partial observations as extra samples.
The sum in \cref{eq:training} becomes $\sum_{t'\in T} \sum_{t\in T_{1:t'}} w_t L\left(Q(s_{t}| O_{\leq t'}),\,a_{t}^{*}\right)$.
For long trajectories, the training data then becomes severely unbalanced between exploration (target not visible) and exploitation (target visible), with a large proportion of the former ($90\%$ of the data in \cref{sec:exploration-experiments}).
We addressed this by reweighting the samples.
We set $w_{t}=\beta^{d_{t}}/\max_{j\in T}\beta^{d_{j}}$ in \cref{eq:training}, \ie a geometric decay scaling with the topological distance to the target $d_t$.
Since expert trajectories are shortest paths, this simplifies to $w_{t}=\beta^{|T|-t}$.

\subsection{Embodied navigation in 3D environments}
Embodied agents such as robots have a pose in 3D space, which for non-holonomic robots limits their available actions (\eg moving forward or rotating), and their observations.

\subsubsection{Embodied pose states (position and orientation)}\label{sec:orientations}
To address the first limitation, we augment the 2D state-space with an extra dimension, which corresponds to $\Theta$ discretised orientations: $\mathcal{S}=\{1,\ldots,N\}^2\times \{1,\ldots,\Theta\}$.
This can be achieved by directly adding one extra dimension to each spatial tensor in \cref{eq:reward_function,eq:proposed_update}.
A table of tensor sizes is in Appendix A.
Note that the much larger state-space makes long-term planning more difficult to learn.
We observed that when naively augmenting the state-space in this way, the models fail to learn correct motion kernels $\Ppredfunc$.
This further reinforces the need for an auxiliary motion loss (\cref{sec:transition-model}), which overcomes this obstacle, and may explain why prior works did not plan through fine-grained rotations.

\subsubsection{3D embeddings for geometric reasoning}\label{sec:3d}
To aggregate image information (CNN embeddings) on to the 2D grid (map) where VIN performs planning, we follow the same strategy as MapNet~\cite{henriques_mapnet_2018}.
Each embedding is associated with a 3D point in world-space via projective geometry, assuming that the camera poses and depths are known (as assumed in prior work \cite{gupta_cognitive_2017,lee_gated_2018,tamar_value_2016,cartillier2021semantic,lenton2021endtoend}, which can be estimated from monocular vision \cite{mur-artal_orb-slam2_2017}).
The embeddings of 3D points \emph{from all past frames} that fall into each cell of the world-space 2D grid are aggregated with mean-pooling.
Due to the use of PointNet~\cite{charles_pointnet_2017} aggregation on cells of a lattice, we named this Lattice PointNet (LPN). 
A self-contained description is in Appendix A.
The LPN has some appealing properties in our context of navigation: 1) it allows reasoning about far away, observed but yet unvisited locations; 2) it fuses multiple observations of the same location, whether from different points-of-view or different times.

\paragraph{Memory-efficient mapping.} 
Temporal aggregation during rollout
can be computed recursively as $e_{t,i,j,k} / n_{t,i,j,k}$ for
$e_{t,i,j,k}=e_{t-1,i,j,k}+e'_{t,i,j,k}$ and $n_{t,i,j,k}=n_{t-1,i,j,k}+n'_{t,i,j,k}$, where $e'_{t,i,j,k}$
is the summed embedding for the points in cell $(i,j,k)$ at time $t$, and $n'_{t,i,j,k}$ is the number of points per cell.
Only the previous map $e_{t-1,i,j,k}$ and previous counts $n_{t-1,i,j,k}$ must be kept, not all past observations.
Thus at run-time the memory cost is \emph{constant} over time, allowing unbounded operation (unlike methods that do not have an explicit map~\cite{savinov_semi-parametric_2018}).

\subsection{Limitations}
Our main contribution is to improve the VIN algorithm itself (sec. \ref{sec:vin}) with correct termination, transition and availability probabilistic models, which is orthogonal to works which build on top of VIN.
We assume the agent's pose, depth image and the camera parameters to be known.
Other dynamic objects in the environment are not modelled.

\section{Experiments}
\label{sec:experiments}

In this section we will gradually build up the capabilities of our method with the proposals from \cref{sec:method}, comparing it to various baselines on increasingly challenging environments, and leading up to unseen real-world 3D environments.

\subsection{2D grid environments}
\label{sec:gridworld}
\begin{figure}
  \centering
  \includegraphics[width=0.325\columnwidth]{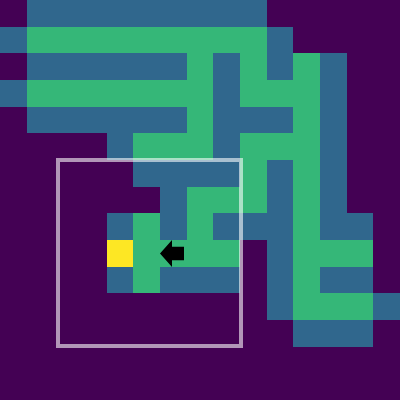}\hfill
  \includegraphics[width=0.325\columnwidth]{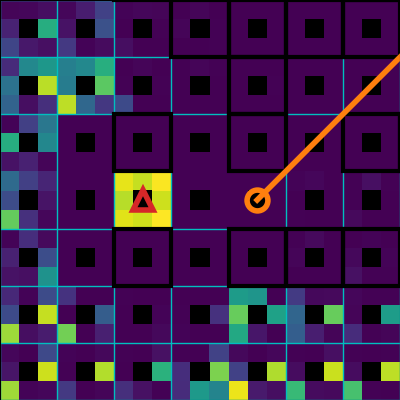}\hfill
  \includegraphics[width=0.325\columnwidth]{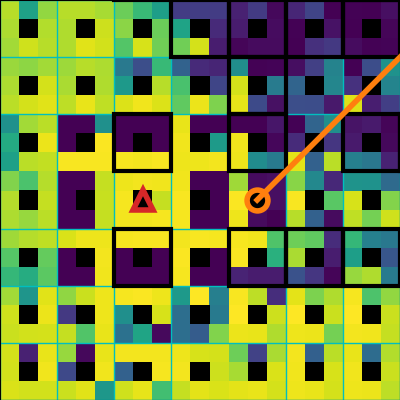}
  \caption{
    Our method on 2D mazes (\cref{sec:embodied-experiments}).
    \textbf{(left)} Input visualisation: unexplored cells are dark, the target is yellow (just found by the agent), and a black arrow shows the agent's position and orientation.
    \textbf{(middle)} Close-up of predicted rewards (higher values are brighter) inside the white rectangle of the left panel. The 3D state-space (position/orientation) is shown, with rewards for the 8 orientations in a radial pattern within each cell (position). Explored cells have low rewards, with the highest reward at the target.
    \textbf{(right)} Close-up of predicted values.
    They are higher facing the direction of the target.
    Obstacles (black border) have low values.
  }
  \label{fig:embodied-run}
\end{figure}

We start with 2D environments, where the observations are top-down views of the whole scene, and which thus do not require dealing with perspective images (\cref{sec:3d}).
Since Tamar \etal~\cite{tamar_value_2016} obtain near-perfect performance in their 2D environments, after reproducing their results we focused on 2D mazes, which are much more challenging since they require frequent backtracking to navigate if exploration is required.
The $15\times 15$ mazes are generated using Wilson's algorithm \cite{wilson_generating_1996}, and an example can be visualised in \cref{fig:embodied-run}.
The allowed moves $\mathcal{A}$ are to any of the 8 neighbours of a cell.
As discussed in \cref{sec:method}, a termination action $D$ must be triggered at the target to successfully complete the task.
The target is placed in a free cell chosen uniformly at random, with a minimum topological distance from the (random) start location equal to the environment size to avoid trivial tasks.

\begin{table}
\caption{Navigation success rate (fraction of trajectories that reach the target) on unseen 2D mazes. Partial observations (exploring an environment gradually) and embodied navigation (translation-rotation state-space) are important yet challenging steps towards full 3D environments.
}
\centering\footnotesize
\setlength{\tabcolsep}{2pt}
\begin{tabular*}{\columnwidth}{@{\extracolsep{\fill}}ccccccc}
\toprule 
 & \multicolumn{3}{c}{Standard loss} & \multicolumn{3}{c}{Reweighted loss (ours)}\tabularnewline
\cmidrule{2-4}\cmidrule{5-7} Env. & VIN & GPPN & CALVIN & VIN & GPPN & CALVIN\tabularnewline
\midrule 
Full obs. & 75.6\tpm{20.6} & 91.3\tpm{8.1} & \textbf{99.0}\tpm{1.0} & 77.5\tpm{26.6} & 96.6\tpm{4.0} & \textbf{99.7}\tpm{0.5} \tabularnewline
\midrule 
Partial & 3.6\tpm{0.6} & 8.5\tpm{3.5} & \textbf{48.0}\tpm{5.2} & 1.7\tpm{1.7} & 11.25\tpm{3.7} & \textbf{92.2}\tpm{1.3} \tabularnewline
\midrule 
Embod. & 11.0\tpm{1.0} & 14.5\tpm{2.1} & \textbf{90.0}\tpm{7.9} & 15.2\tpm{3.6} & 28.5\tpm{3.5} & \textbf{93.7}\tpm{6.2} \tabularnewline
\bottomrule
\end{tabular*}\label{tab:2d}
\end{table}

\subsubsection{Fully-known environment with positional states}
\label{sec:fully-known}
\paragraph{Baselines and training.}
For the first experiment, we compare our method (CALVIN) with other differentiable planners: the VIN~\cite{tamar_value_2016} and the more recent GPPN~\cite{lee_gated_2018}, on fully-observed environments.
Other than using mazes instead of convex obstacles, this setting is close to Tamar \etal's~\cite{tamar_value_2016}.
The VIN, GPPN and CALVIN all use 2-layer CNNs to predict their inputs (details in Appendix A).
All networks are trained with $4K$ example trajectories in an equal number of different mazes, using the Adam optimiser with the optimal learning rate chosen from $\{0.01, 0.005,0.001\}$, until convergence (up to 30 epochs). Navigation success rates (fraction of trajectories that reach the target) for epochs with minimum validation loss are reported.
Our reweighted loss (\cref{sec:reweighting}) is equally applicable to all differentiable planners, so we report results both with and without it.

\paragraph{Results.}
\Cref{tab:2d} (first row) shows the navigation success rate, averaged over 3 random seeds (and the standard deviation).
The VIN has a low success rate, showing that it does not scale to large mazes.
GPPN achieves a high success rate, and CALVIN performs near-perfectly.
This may be explained by the GPPN's higher capacity, as it contains a LSTM with more parameters.
Nevertheless, CALVIN has a more constrained architecture, so its higher performance hints at a better inductive bias for navigation.
It is interesting to note that the proposed reweighted loss has a beneficial effect on all methods, not just CALVIN.
With the correct data distribution, any method with sufficient capacity can fit the objective.
This shows that addressing the unbalanced nature of the data is an important, complementary factor.

\subsubsection{Partially observable environment}\label{sec:exploration-experiments}
\paragraph{Setting.}
Next, we compare the same methods in unknown environments, where the observation maps only contain observed features up to the current time step (sec.~\ref{sec:exploration}).
To simulate local observations, we perform ray-casting to identify cells that are visible from the current position, up to 2 cells away.
Example observations are in Appendix A.

\paragraph{Results.}
In this case, the agent has to take significantly more steps to explore compared to a direct route to the target.
From \cref{tab:2d} (2\textsuperscript{nd} row) we can see that partial observability causes most methods to fail catastrophically.
The sole exception is CALVIN with the reweighted loss (proposed), which performs well.
Note that to succeed, an agent must acquire several complex behaviours: directing exploration to large unseen areas; backtracking from dead ends; 
and seeking the target when seen.
Our method displays all of these behaviours (see Appendix A for visualisations).
While the model initially assigns high values to all unexplored states, as soon as the target is in view, the model assigns a high value to the target state and its neighbours.
Since only a combination of CALVIN and a reweighted loss works at all, we infer that a correct inductive bias and a balanced data distribution are both necessary for success.

\subsubsection{Embodied navigation with orientation}\label{sec:embodied-experiments}
\paragraph{Setting.}
Now we consider embodied navigation, where transitions depend on the agent's orientation (sec.~\ref{sec:orientations}).
We augment the state-space of all methods (\cref{sec:orientations}) with 8 orientations at $\ang{45}$ intervals, and allow 4 move actions $\mathcal{A}$: forward, backward, and rotating in either direction.

\paragraph{Results.}
In \cref{tab:2d} (3\textsuperscript{rd} row) we observe that VIN and GPPN perform slightly better, but still have a low chance of success. CALVIN outperforms them by a large margin.
We also visualise a typical run in \cref{fig:embodied-run} (refer to the caption for more detailed analysis).
One advantage of CALVIN displayed in \cref{fig:values,fig:embodied-run} is that values and rewards are fully interpretable and play the expected roles in value iteration (\cref{eq:vi-grid}).
Less constrained architectures~\cite{tamar_value_2016,lee_gated_2018} insert operators that deviate from the value iteration formulation, and thus lose their interpretability as rewards and values (\cf Appendix A).

\begin{table}
\caption{Navigation success rate on unseen 3D mazes (MiniWorld).
Note that the baselines do not generalise to larger mazes.
}
\centering\footnotesize
\begin{tabular*}{\columnwidth}{@{\extracolsep{\fill}}ccccccc}
\toprule 
 & \multicolumn{2}{c}{CNN backbone} & \multicolumn{3}{c}{LPN backbone (ours)}\\
\cmidrule{2-3}\cmidrule{4-6} Size & A2C & PPO & VIN & GPPN & CALVIN (ours)\\
\midrule 
$3\times 3$ & \textbf{98.7}\tpm{1.9} & 81.0\tpm{26.9} & 90.3\tpm{3.1} & 91.3\tpm{4.7} & \textbf{97.7}\tpm{1.7} \\
\midrule 
$8\times 8$ & 23.6\tpm{4.9} & 14.7\tpm{6.2} & 41.2\tpm{9.5} & 33.3\tpm{8.6} & \textbf{69.2}\tpm{5.3} \\
\bottomrule
\end{tabular*}\label{tab:miniworld}
\end{table}

\subsection{3D environments}
Having validated embodied navigation and exploration, we now integrate the Lattice PointNet (LPN, \cref{sec:3d}) to handle first-person views of 3D environments.

\subsubsection{Synthetically-rendered environments}\label{sec:miniworld}
\begin{figure}
  \centering
  \renewcommand*{\arraystretch}{0}
  \begin{tabular}{*{3}{@{}c}@{}}
  \includegraphics[height=2.4cm]{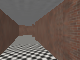} &
  \includegraphics[height=2.4cm]{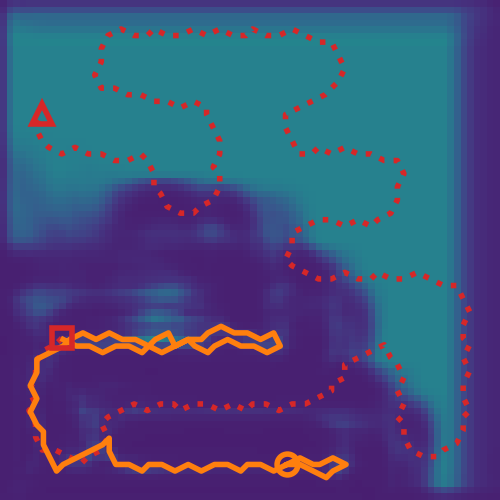} &
  \includegraphics[height=2.4cm]{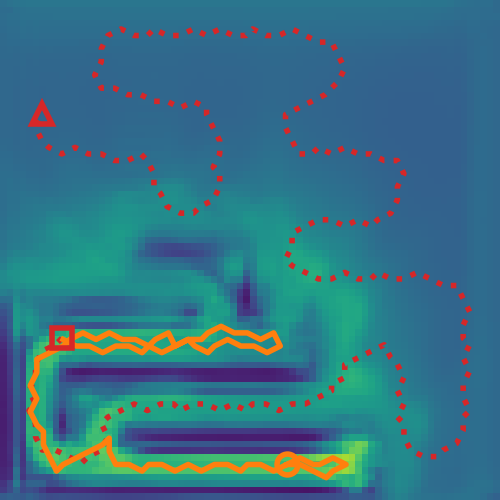} \\
  \includegraphics[height=2.4cm]{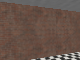} &
  \includegraphics[height=2.4cm]{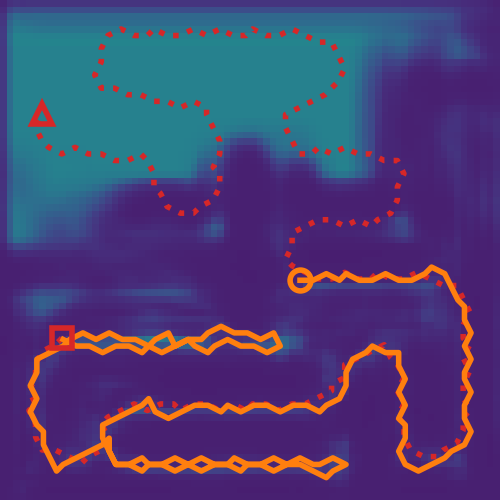} &
  \includegraphics[height=2.4cm]{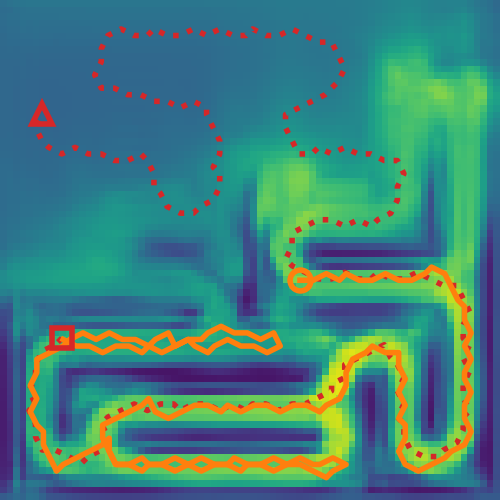} \\
  \includegraphics[height=2.4cm]{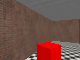} &
  \includegraphics[height=2.4cm]{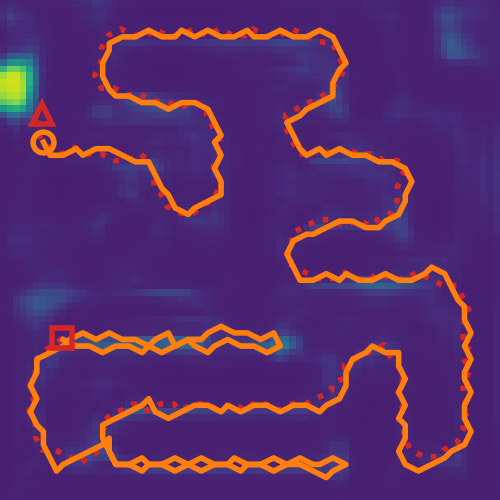} &
  \includegraphics[height=2.4cm]{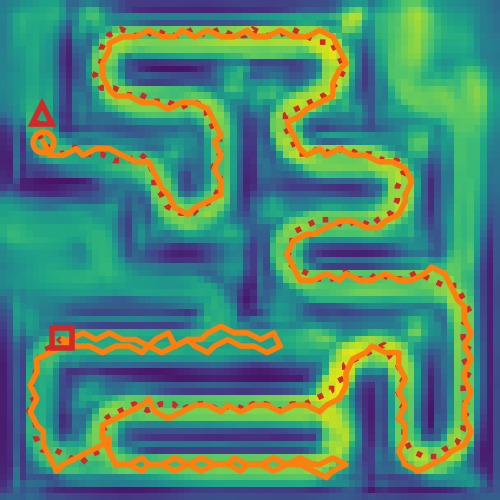}
  \end{tabular}
  \caption{
    Example results on MiniWorld (\cref{sec:miniworld}). Left to right: input images, predicted rewards and values. The format is as in \cref{fig:avd}. Notice the high reward on unexplored regions, replaced with a single peak around the target when it is seen (last row).
  }
  \label{fig:miniworld}
\end{figure}

\paragraph{Dataset.}
We used the MiniWorld simulator~\cite{gym_miniworld}, which allowed us to easily generate 3D maze environments with arbitrary layouts.
Only a monocular camera is considered (not \ang{360} views~\cite{lee_gated_2018}).
The training trajectories now consist of first-person videos of the shortest path to the target, visualised in \cref{fig:miniworld}.
We randomly generate 1K trajectories in mazes on either small ($3\times 3$) or large ($8\times 8$) grids by adding or removing walls at the boundaries of this grid's cells.
Note that the maze's layout and the agent's location do not necessarily align with the 2D grids used by the planners (as in \cite{lee_gated_2018}).
Thus, planning happens on a fine discretisation of the state-space ($30\times30$ for small mazes and $80\times80$ for large ones, with $8$ orientations). This allows smooth motions and no privileged information about the environment.
Both translation and rotation are perturbed by Gaussian noise, forcing all agents to model uncertain dynamics. 

\paragraph{Baselines and training.}
We compare several baselines: two popular RL methods, A2C~\cite{mnih_asynchronous_2016} and PPO~\cite{Schulman2017ProximalPO}, as well as the VIN, GPPN and the proposed CALVIN.
Since A2C and PPO are difficult to train if triggering the ``done'' action is strictly required, we relaxed the assumption, allowing the agent to terminate once it is in close proximity to the target. 
All methods use as a first stage a simple 2-layer CNN (details in Appendix A).
Since this CNN was not enough to get the VIN, GPPN and CALVIN to work well (see Appendix A), they all use our proposed LPN backbone.
We could not include GPPN's strategy of taking as input views at all possible states~\cite{lee_gated_2018}, due to the high memory requirements and the environment not being fully visible (only local views are available).
Other training details are identical to \cref{sec:gridworld}.

\paragraph{Results.}
In \cref{tab:miniworld} we observe that, although the RL methods are successful in small mazes, their reactive policies cannot scale to large mazes.
CALVIN succeeds reliably even for longer trajectories, outperforming the others.
It is interesting to note that the proposed LPN backbone is important for all differentiable planners, and it allows them to achieve very high success rates for small environments (though our CALVIN method performs slightly better).
We show an example run of our method in \cref{fig:miniworld}, where it found the target after an efficient exploration period, despite not knowing its location and never encountering this maze before.

\begin{figure}[b]
  \centering
  \includegraphics[width=0.12\columnwidth]{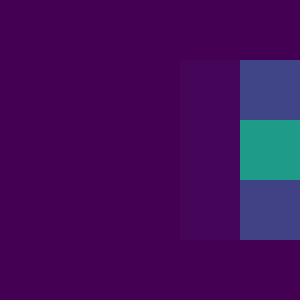}\hfill
  \includegraphics[width=0.12\columnwidth]{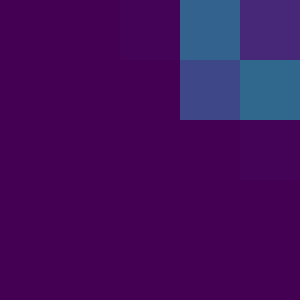}\hfill
  \includegraphics[width=0.12\columnwidth]{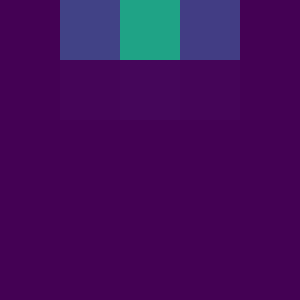}\hfill
  \includegraphics[width=0.12\columnwidth]{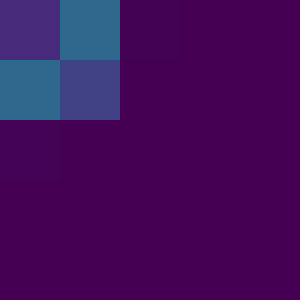}\hfill
  \includegraphics[width=0.12\columnwidth]{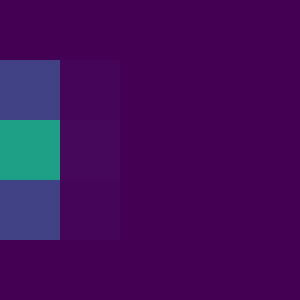}\hfill
  \includegraphics[width=0.12\columnwidth]{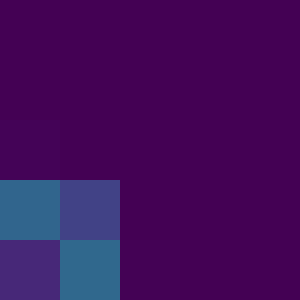}\hfill
  \includegraphics[width=0.12\columnwidth]{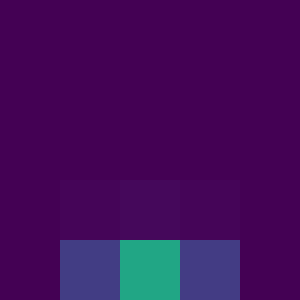}\hfill
  \includegraphics[width=0.12\columnwidth]{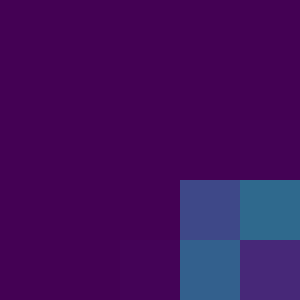}
  \caption{
    Transition model learnt from MiniWorld trajectories for the \textit{move forward} action at each discretised orientation, at $\ang{45}$ intervals. Higher values are brighter, and lower values are darker (purple for a probability of 0).
  }
  \label{fig:motion_models}
\end{figure}

\Cref{fig:motion_models} visualises the learnt state transitions $\Ppredfunc$ for the \textit{move forward} action in CALVIN. We observe that the learning mechanism outlined in \cref{sec:transition-model} works even for transitions with added noise, which is the case for MiniWorld experiments. The network learns to propagate values probabilistically from the possible next states.

\subsubsection{Indoor images from a real-world robot}\label{sec:avd}
Finally, we tested our method on real images obtained with a robotic platform.

\paragraph{Dataset.}
We used the Active Vision Dataset~\cite{ammirato2017dataset} (AVD), which allows interactive navigation with real image streams, without synthetic rendering (as opposed to~\cite{habitat19iccv}).
This is achieved using monocular RGBD images from 19 indoor environments, densely collected by a robot on a 30cm grid and at \ang{30} rotations.
The set of over 30K images can be composed to simulate any trajectory, up to the some spatial granularity.
There are also bounding box annotations of object instances, which we use to evaluate semantic navigation.
We use the last four environments as a validation set, and the rest for training (by sampling 1K shortest paths to the target).
A visualisation is shown in \cref{fig:avd}.

\paragraph{Tasks and training.}
We considered a semantic navigation task of seeking an object of a learnt class. We chose the most common class (``soda bottle'') as a target object.
Training follows \cref{sec:miniworld}.

\paragraph{Results.}
We report the performances of VIN, GPPN and CALVIN after 8 epochs of training in \cref{tab:avd}.
As in \cref{sec:miniworld}, they also fail without the LPN, so all results are with the LPN backbone.
A similar conclusion to that for synthetic environments can be drawn: proper spatio-temporal aggregation of local observations is essential for differentiable planners to scale realistically.
CALVIN achieves a significantly higher success rate on the training set than the other methods.
On the other hand, while it has higher mean validation success, the high variance of this estimate does not allow the result to be as conclusive as for training. We can attribute this to the small size of AVD in general, and of the validation set in particular, which contains only 3 different indoor scenes.
Nevertheless, this shows that CALVIN learns effective generic strategies to seek a specific object, and that VIN and GPPN equipped with a similar backbone can achieve partial success in several training environments.

\begin{table}
\caption{Navigation success rate on the Active Vision Dataset, with
real robot images taken in indoor spaces. The task is to navigate
to an object of a learned class.
All methods use the proposed LPN backbone, as they fail without it.}
\centering\footnotesize
\setlength{\tabcolsep}{2pt}
\begin{tabular*}{0.85\columnwidth}{@{\extracolsep{\fill}}cccc}
\toprule 
Subset & VIN & GPPN & CALVIN (ours)\tabularnewline
\midrule 
Training & 61.6\tpm{4.5} & 50.6\tpm{9.2} & \textbf{70.3}\tpm{4.9} \tabularnewline
\midrule 
Validation & 45.0\tpm{1.0} & 44.0\tpm{3.5} & \textbf{47.6}\tpm{6.0} \tabularnewline
\bottomrule
\end{tabular*}\label{tab:avd}
\end{table}

\section{Discussion}
\label{sec:conclusion}
We analysed several shortcomings of current differentiable planners, with the goal of deploying to real robot platforms.
We found that they can be addressed by highly complementary solutions: a constrained transition model (CALVIN) correctly incorporates illegal actions (as opposed to simply discouraging them via rewards) and task termination; a 3D state-space accounts for robot orientation; a LPN backbone efficiently fuses spatio-temporal information; and trajectory reweighting addresses unbalanced training data.
We provide empirical evidence in several settings,
including on images of unseen environments from a real robot.

Although deep networks hold promise for making navigation more robust under uncertainty, their uninterpretable failure modes mean that they are not yet mature enough for safety-critical applications, and more research to close this gap is still needed.
Vehicle applications are especially important, due to the high masses and velocities involved.
We propose a more interpretable MDP structure for VINs, and train our model solely with safe offline demonstrations.
However, their reliability is far from guaranteed, and complementary safety systems in hardware must be considered in any deployment.
In addition to further improving robustness to failures, future work may investigate more complex tasks, such as natural language commands, and study the effect of sensor drift on navigation performance.

\paragraph{Acknowledgements.} The authors would like to acknowledge the generous support of the Royal Academy of Engineering (RF\textbackslash 201819\textbackslash 18\textbackslash 163) and the Ezoe Memorial Recruit Foundation.
They would also like to thank the authors of MiniWorld~\cite{gym_miniworld} and the Active Vision Dataset~\cite{ammirato2017dataset} for making their source code and data available.

{\small
\bibliographystyle{ieee_fullname}
\bibliography{egbib, references}
}

\newpage

\appendix

\section{}

\subsection{Implicit assumptions of the VIN architecture}

\paragraph{Differences between VIN and the idealised VI on a grid.}
Comparing eq. 1 in sec. 3 to eq. 2 in sec. 3.1, we note several differences:

\noindent
1. The VIN value estimate $V$ takes the maximum action-value $Q$ across \emph{all possible actions} $\mathcal{A}$, even illegal ones (e.g. moving into an obstacle).
Similarly, the $Q$ estimate is also updated even for illegal actions.
In contrast, VI only considers legal actions for each state (cell), i.e. $\mathcal{A}(i,j)$.

\noindent
2. The VIN reward $\widehat{R}$ is assumed independent of the action.
This means that, for example, a transition between two states cannot be penalised directly; a penalty must be assigned to one of the states (regardless of the action taken to enter it).

\noindent
3. The VIN transition probability is expanded into 2 terms, $P^R$ which affects the reward and $P^V$ which affects the estimated values.
This decoupling means that they do not enjoy the physical interpretability of VI's $P$ (i.e. probability of state transitions), and rewards and values can undergo very different transition dynamics.

\noindent
4. Unlike the VI, the VIN considers the state transition translation-invariant.
This means that it cannot model obstacles (illegal transitions) using $P$, and must rely on assigning a high penalty to the reward $\widehat{R}$ of those states instead.

\subsection{Implementation details}
\subsubsection{Embodied pose states}
One of our contributions is extending the VIN framework to accommodate embodied pose states, i.e. states which encode both position and orientation. 
We achieve this by augmenting the 2D state-space with an extra dimension for orientations.
\Cref{table:embodied} shows correspondences between tensor dimensions of the positional method and the embodied method for each component of the architecture. $X$ and $Y$ are the size of the internal spatial discretisation of the environment, $M$ is the internal discretisation of the orientation, $A$ is the number of actions, and $K$ is the kernel dimension for spatial locality. 

Note that the value iteration step in \methodname{} performs a 2D convolution of $\widehat{P}$ over a 2D value map in the case of positional states and a 3D convolution over a 3D value map with orientation in the case of embodied pose states. In the embodied case, the second dimension of $\widehat{P}$ corresponds to the orientation of the current state, and the third dimension corresponds to that of the next state.

\begin{table}[h]
\centering
\caption{Comparison of individual components in the implementation of \methodname{} for positional states and for embodied pose states.}
\begin{tabular}{c | l l}
 \hline
 & Positional & Embodied \\ [0.5ex] 
 \hline
 State $s$ & $(x,y)$ & $(\theta, x, y)$ \\
 VI step & Conv2d & Conv3d \\
 \hline
 $V(s)$ & $X \times Y$ & $M \times X \times Y$ \\
 $Q(s,a)$ & $A \times X \times Y$ & $A \times M \times X \times Y$ \\
 $\Apredfunc$ & $A \times X \times Y$ & $A \times M \times X \times Y$ \\
 $\Ppred$ & $A \times K \times K$ & $A \times M \times M \times K \times K$ \\
 $\Rpred$ & $A \times K \times K$ & $A \times M \times M \times K \times K$ \\ [1ex] 
 \hline
\end{tabular}
\label{table:embodied}
\end{table}

\subsubsection{3D embeddings for geometric reasoning}\label{sec:lpn}

Since the learnable functions ($\Ppred$, $\Apred$ and $\Rpred$) in our proposed method (and other VIN-based methods) are 2D CNNs, their natural input is a 2D grid of $m$-dimensional embeddings, denoted $e_{tij}\in\mathbb{R}^{m}$, for time $t$ and discrete world-space coordinates $(i,j)$. This can be interpreted as a spatio-temporal map tensor. We then wish to project and aggregate useful semantic information from an image $I_{t}$, extracted by a CNN $\phi$, into this tensor. This requires both knowledge of the camera position $c_{t}$ and rotation matrix $R_{t}$, which we assume following previous work \cite{gupta_cognitive_2017,lee_gated_2018,tamar_value_2016} (and which can be estimated from monocular vision \cite{mur-artal_orb-slam2_2017}). Spatial projection also requires knowing (or estimating) the depths $d_{t}(p)$ of each pixel $p$ in $I_{t}$ (either with a RGBD camera as in our experiments, or monocular depth estimation~\cite{fu2018deep}). We can then write the homogenous 3D coordinates of each pixel $p$ in the absolute reference frame using projective geometry~\cite{hartley2003multiple}:
\begin{equation}
\left[x_{t}(p),\,y_{t}(p),\,z_{t}(p),\,1\right]=c_{t}+R_{t}K\left[p_{1},p_{2},d_{t}(p),1\right]^{\top},
\end{equation}
where $K$ is the camera's intrinsics matrix. Given these absolute coordinates of pixel $p$, we can calculate the closest map embedding $e_{tij}$ to it, and thus aggregate the CNN embeddings $\phi\left(I_{t}\right)$ associated with all pixels close to a map cell. Inspired by PointNet~\cite{charles_pointnet_2017}, we choose mean-pooling for aggregation. Since we have spatial aggregation, we can easily extend this framework to work spatio-temporally, aggregating information from past frames $t'\leq t$. More formally:
\begin{alignat}{1}
e_{tij} & =\mathrm{avg}_{t'\leq t}\left\{ \phi_{p}\left(I_{t'}\right):\,\tau i\leq x_{t'}(p)<\tau(i+1),\right. \notag\\
 & \qquad\quad\;\;\left.\tau j\leq y_{t'}(p)<\tau(j+1),\right. \notag\\
 & \qquad\quad\;\;\left.\tau k\leq z_{t'}(p)<\tau(k+1),p\in I_{t'}\right\} \label{eq:lpn}
\end{alignat}
where $\tau$ is the absolute size of each square grid cell,
avg averages the elements of a set,
and $\phi_{p}\left(I_{t'}\right)$ retrieves the CNN embedding of image $I_{t'}$ for pixel $p$. Due to the similarity between \cref{eq:lpn} and a PointNet embedded on a 2D lattice, we named it Lattice PointNet (LPN). Other than the lattice embedding, there are other major differences from the PointNet: we apply it spatio-temporally with a causal constraint ($t'\leq t$), and the downstream predictors that take it as input ($\Ppred(e_{t})$, $\Apred(e_{t})$ and $\Rpred(e_{t})$) are 2D CNNs that can reason spatially in the lattice, as opposed to the PointNet's unstructured multi-layer perceptrons~\cite{charles_pointnet_2017}. A related proposal for SLAM used spatial max-pooling but more complex LSTMs/GRUs for temporal aggregation \cite{henriques_mapnet_2018,cartillier2021semantic}. Another related work on end-to-end trainable spatial embeddings uses egospherical memory \cite{lenton2021endtoend}.

\subsubsection{Architectural design of Lattice PointNet}

The Lattice PointNet described in \cref{sec:lpn} consists of three stages: a CNN that extracts embeddings from observations in image-space (image encoder), a spatial aggregation step (eq. 10 in sec. 4.2.2) that performs mean pooling of embeddings for each map cell, and another CNN that refines the map embedding (map encoder). The image encoder consists of two CNN blocks, each consisting of the following layers in order: optional group normalisation, 2D convolution, dropout, ReLU and 2D max pooling. The map encoder consists of 2D convolution, dropout, ReLU, optional group normalisation, and finally, another 2D convolution. The number of channels of each convolutional layer are $(80, 80, 80, 40)$ for MiniWorld and $(40, 40, 40, 20)$ for AVD respectively.
The point clouds can consume a significant amount of memory for long trajectories. Hence, we use the most recent $40$ frames for the $8\times8$ MiniWorld maze.

The input to the LPN is a 3-channel RGB image for the MiniWorld experiment, and a 128-channel embedding extracted using the first 2 blocks of ResNet18 pre-trained on ImageNet for the AVD experiment.

\subsubsection{Architectural design of the CNN backbone}
This CNN backbone is used in a control experiment in \cref{sec:cnn_backbone} to show the effectiveness of the LPN backbone. In contrast to LPN which performs spatial aggregation of embeddings, the CNN backbone is a direct application of an encoder-decoder architecture that transforms image-space observations into map-space embeddings. Gupta \etal \cite{gupta_cognitive_2017} employed a similar architecture to obtain their map embeddings. While they use ResNet50 as the encoder network, we used a simple CNN for the MiniWorld experiment to match the result obtained with LPN. 

The CNN backbone consists of three stages: a CNN encoder, two fully-connected layers with ReLU to transform embeddings from image-space to map-space, and a CNN decoder. The encoder consists of 3 blocks of batch normalisation, 2D convolution, dropout, ReLU and 2D max pooling, and a final block with just batch normalisation and 2D convolution. The number of channels of each convolutional layer are $(64, 128, 128, 128)$, respectively. 

The fully-connected layers take in an input size of $128 \times 5 \times 7$, reduces it to a hidden size of $128$, and outputs either $128 \times 5 \times 5$ for the smaller maze or $128 \times 4 \times 4$ for the larger maze, which is then passed to the decoder.

The decoder consists of 3 blocks of batch normalisation, 2D deconvolution, dropout and ReLU, and a final block with just 2D deconvolution. The number of channels of each deconvolution layers are $(128, 128, 64, 20)$, respectively.
The output size of the decoder depends on the map resolution, hence we chose appropriate strides, kernel sizes and paddings in the decoder network to match the output sizes of $30 \times 30$ and $80 \times 80$. This approach is not scalable to maps with high resolution or with arbitrary size, which is one of the drawbacks of this approach. 

\subsection{Experiment setup}

\subsubsection{Expert trajectory generation}
Expert trajectories are generated by running an A*~\cite{a_star} planner from the start state to the target state. We assigned Euclidean costs to every transition in the 2D grid environments, and a cost of $1$ per move for the MiniWorld and AVD environments. In the case of MiniWorld, an additional cost is assigned to locations near obstacles to ensure that the trajectories are not in close proximity to the walls. 

\subsubsection{Hyperparameter choices}
Similarly to VIN~\cite{tamar_value_2016} which uses a 2-layer CNN to predict the reward map, and GPPN~\cite{lee_gated_2018}, which uses a 2-layer CNN to produce inputs to the LSTM, CALVIN uses a 2-layer CNN as an available actions predictor $\Apredfunc$. For each experiment, we chose the size of the hidden layer from $\{40, 80, 150\}$. $150$ was used for all the grid environments, $80$ for MiniWorld and $40$ for AVD, partially due to memory constraints. 

VIN has an additional hyperparameter for the number of hidden action channels, which we set to $40$, which is sufficiently bigger than the number of actual actions in all of our experiments.
While the kernel size $K$ for VIN and CALVIN were set to $3$ for experiments in the grid environment, it was noted in \cite{lee_gated_2018} that GPPN works better with larger kernel size. Therefore, we chose the best kernel size out of $\{3, 5, 7, 9, 11\}$ for GPPN. For experiments on MiniWorld and AVD, there are state transitions with step size of $2$, hence we chose $K=5$ for VIN and CALVIN.

The number of value iteration steps $k$ was chosen from $\{20, 40, 60, 80, 100\}$. For trajectory reweighting, $\beta$ was chosen from $\{0.1, 0.25, 0.5, 0.75, 1.0\}$. 

\subsubsection{Rollout at test time}
We test the performance of the model by running navigation trials (rollouts) on a randomly generated environment. At every time step, the model is queried the set of $Q$-values $\{Q(s, a): a \in \mathcal{A}\}$ for the current state $s$, and an action which gives the maximum predicted $Q$ value is taken.

While VIN is trained with $V^{(0)}$ initialised with zeros, in a true Value Iteration algorithm, the value function must converge for an optimal policy to be obtained.
To help the value function converge faster under a time and compute budget, we initialise the value function with predicted values from the previous time step at test time with online navigation.

We set a limit to the maximum number of steps taken by the agent, which were $200$ for the fully-known $15 \times 15$ grid, $500$ for the partially known grid, $300$ for MiniWorld $(3 \times 3)$, $1000$ for MiniWorld $(8 \times 8)$, and $100$ for AVD.

\subsection{Additional experiments}

\subsubsection{Ablation study of removing loss components}
CALVIN is trained on three additive loss components: a loss term for the predicted Q-values $L_Q$ (sec.4.1.1), a loss term for the transition models $L_P$ (sec.4.1.2), and a loss term for the action availability $L_A$ (sec.4.1.3). We assessed the contribution of each loss component to the overall performance.

We conducted the experiments on the partially observable grid environment (sec. 5.1.2). The results in \cref{tab:ablation} indicate that all loss components, in particular the transition model loss, contributes to the robust performance of the network.

\begin{table}[h]
\caption{Navigation success rate of CALVIN in partially observable 2D mazes with loss components removed.
}
\centering\footnotesize
\setlength{\tabcolsep}{2pt}
\begin{tabular*}{\columnwidth}{@{\extracolsep{\fill}}cccc}
\toprule 
Loss & $L_Q + L_P + L_A$ & $L_Q + L_P$ & $L_Q + L_A$ \tabularnewline
\midrule 
Success rate & 92.2 & 84.1 & 8.3\tabularnewline
\bottomrule
\end{tabular*}\label{tab:ablation}
\end{table}

\subsubsection{Comparison of LPN against CNN backbone}
\label{sec:cnn_backbone}

We compared our proposed LPN backbone against a typical encoder-decoder CNN backbone as a component that maps observations to map embeddings. We evaluated the performance of the two methods for VIN, GPPN and CALVIN. In \cref{tab:miniworld_appendix}, we observe that LPN backbone is highly effective, especially for larger environments where long-term planning based on spatially aggregated embeddings is necessary.

\begin{table}[h]
\caption{Navigation success rate on unseen 3D mazes (MiniWorld).
Most methods do not generalise to larger mazes.
The proposed LPN demonstrates robust performance in larger unseen mazes. 
}
\centering\footnotesize
\begin{tabular*}{\columnwidth}{@{\extracolsep{\fill}}ccccccc}
\toprule 
 & \multicolumn{3}{c}{CNN backbone} & \multicolumn{3}{c}{LPN backbone (ours)}\\
\cmidrule{2-4}\cmidrule{5-7} Size & VIN & GPPN & CALVIN & VIN & GPPN & CALVIN\\
\midrule 
$3\times 3$ & 89.4 & 73.1 & 75.2 & 90.3 & 91.3 & \textbf{97.7} \\
\midrule 
$8\times 8$ & 0.6 & 18.3 & 8.6 & 41.2 & 33.3 & \textbf{69.2} \\
\bottomrule
\end{tabular*}\label{tab:miniworld_appendix}
\end{table}

\begin{figure}
  \centering
  \includegraphics[width=0.33\columnwidth]{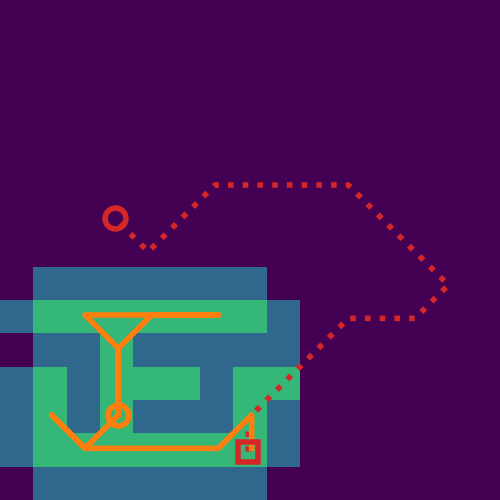}\hfill
  \includegraphics[width=0.33\columnwidth]{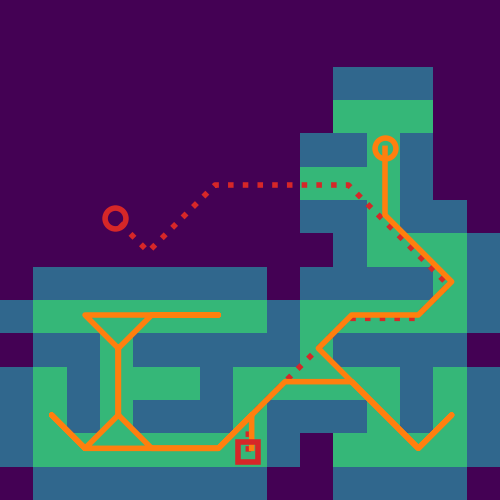}\hfill
  \includegraphics[width=0.33\columnwidth]{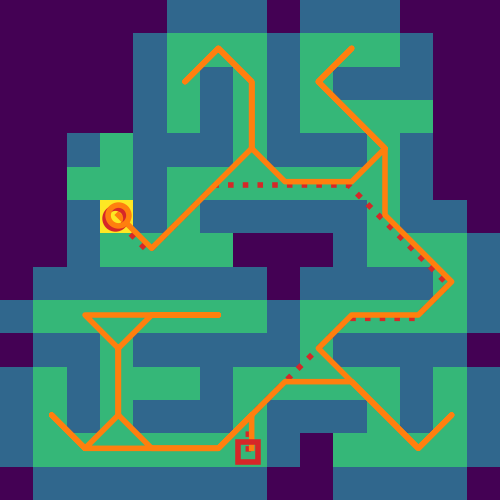}
  \includegraphics[width=0.33\columnwidth]{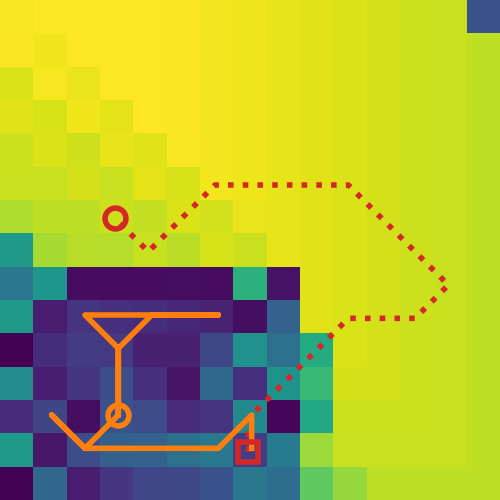}\hfill
  \includegraphics[width=0.33\columnwidth]{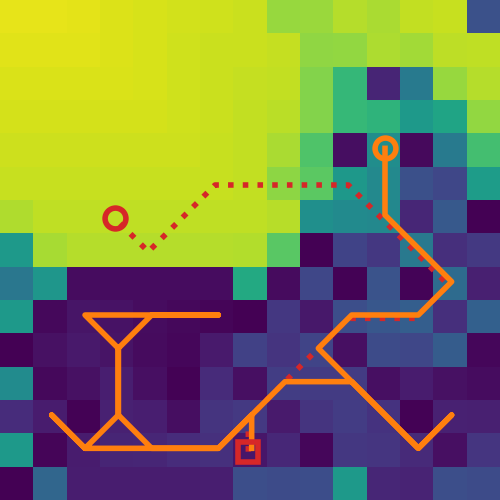}\hfill
  \includegraphics[width=0.33\columnwidth]{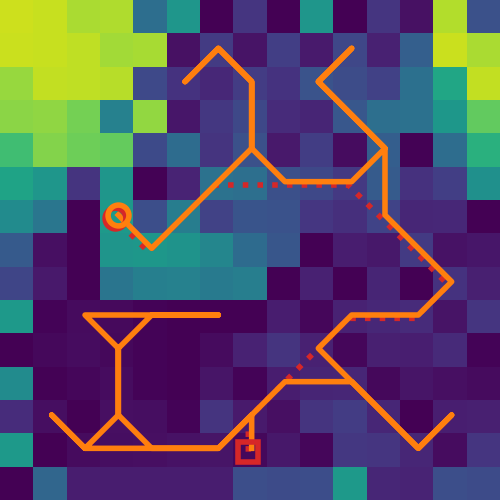}
  \caption{
    Example rollout of CALVIN after $21$ steps (left column), $43$ steps (middle column) and $65$ steps (right column). CALVIN successfully terminated at $65$ steps.
    \textbf{(top row)} Input visualisation: unexplored cells are dark, the discovered target is yellow. The correct trajectory is dashed, the current one is solid. The orange circle shows the position of the agent.
    \textbf{(bottom row)} Predicted values (higher values are brighter). Explored cells have low values, while unexplored cells and the discovered target are assigned high values. 
  }
  \label{fig:calvin_rollout}
\end{figure}

\subsection{Example rollout in a partially observable maze}
\label{sec:calvin_rollout}
We present an example of a trajectory taken by CALVIN at runtime, with corresponding observation maps and predicted values in \Cref{fig:calvin_rollout}. At each rollout step, \methodname{} performs inference on the best action to take based on its current observation map.
No information about the location of the target is given until it is within view of the agent.
This makes the problem challenging, since the agent may have to take significantly more steps compared to an optimal route to reach the target.
In this example, the agent managed to backtrack every time it encountered a dead end, successfully reaching the target after 65 steps. 
The model initially assigns high values to all unexplored states. When the target comes into view, the model assigns a high probability to the availability of the ``done'' action at the corresponding state. The agent learns a sufficiently high reward for a successful termination so that the ``done'' action is triggered at the target.

\begin{figure}
  \centering
  \includegraphics[width=0.33\columnwidth]{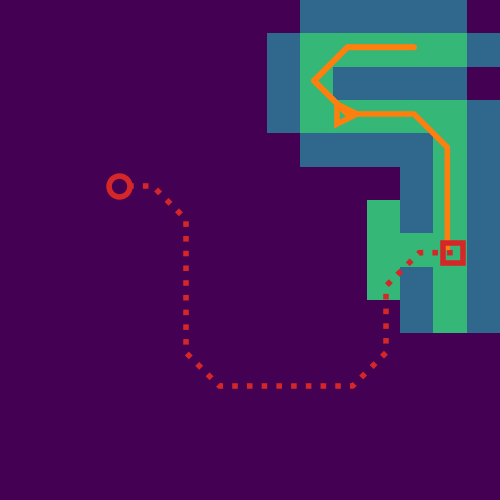}\hfill
  \includegraphics[width=0.33\columnwidth]{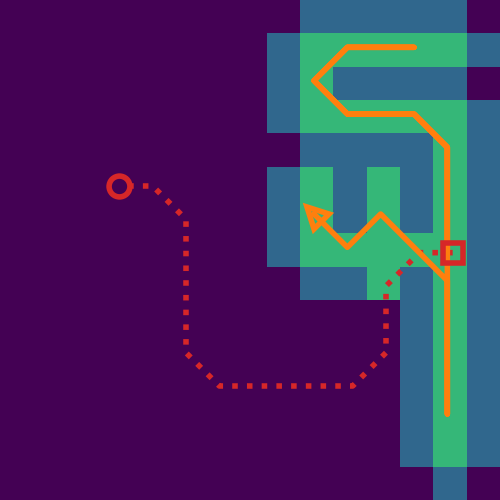}\hfill
  \includegraphics[width=0.33\columnwidth]{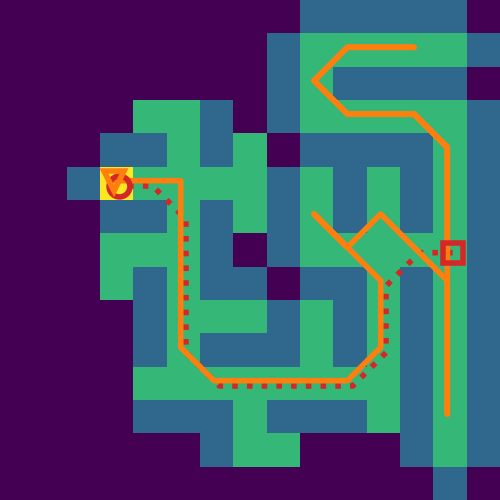}
  \includegraphics[width=0.33\columnwidth]{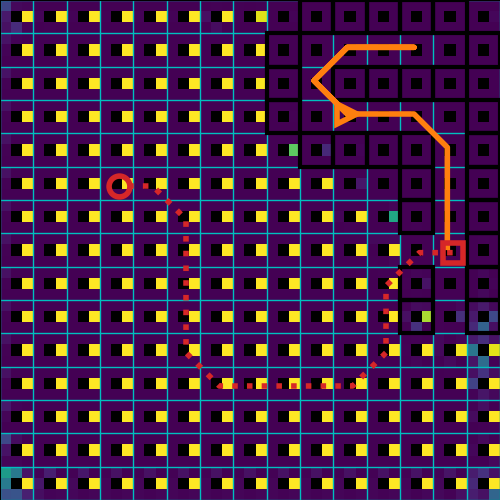}\hfill
  \includegraphics[width=0.33\columnwidth]{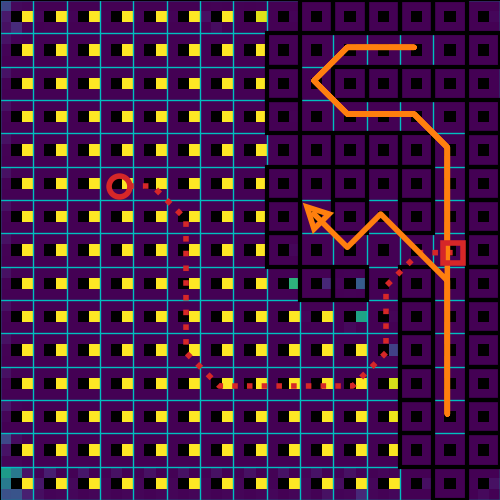}\hfill
  \includegraphics[width=0.33\columnwidth]{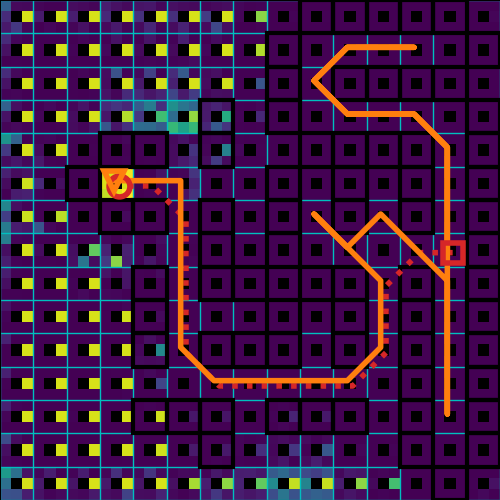}
  \includegraphics[width=0.33\columnwidth]{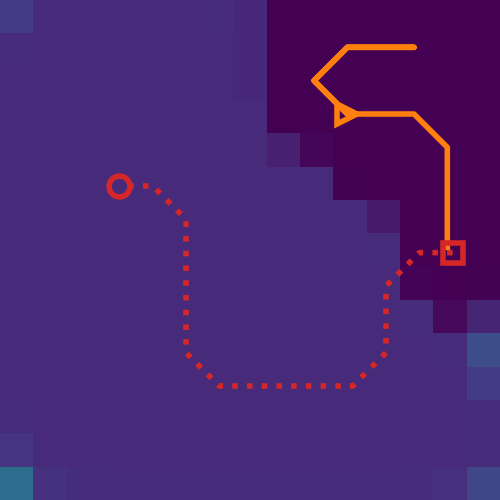}\hfill
  \includegraphics[width=0.33\columnwidth]{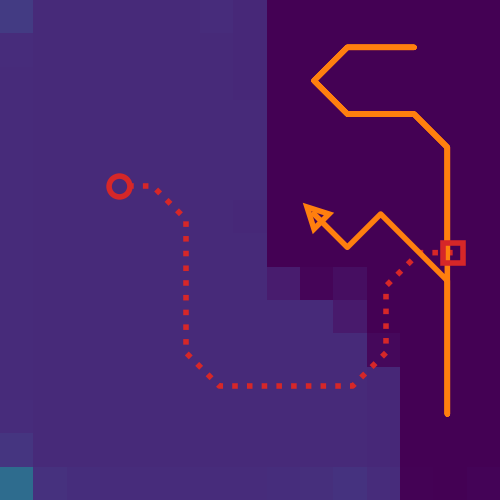}\hfill
  \includegraphics[width=0.33\columnwidth]{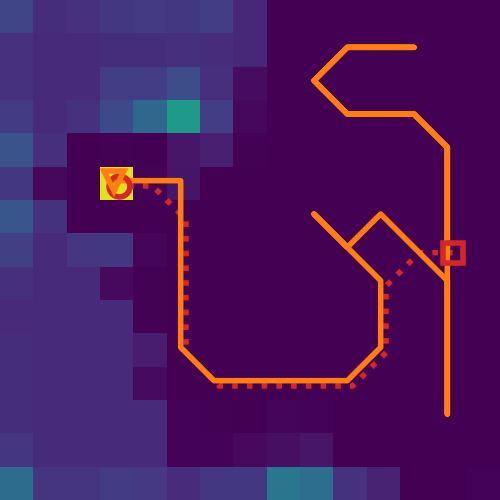}
  \includegraphics[width=0.33\columnwidth]{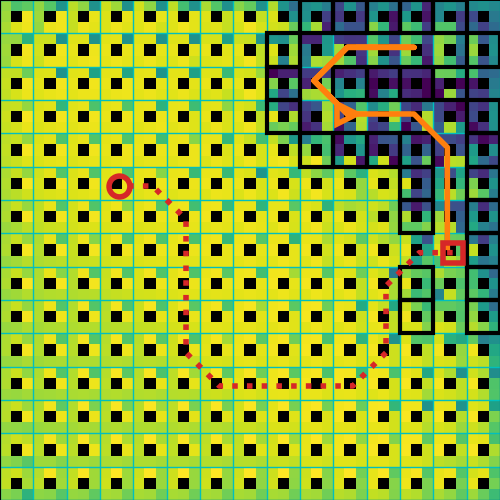}\hfill
  \includegraphics[width=0.33\columnwidth]{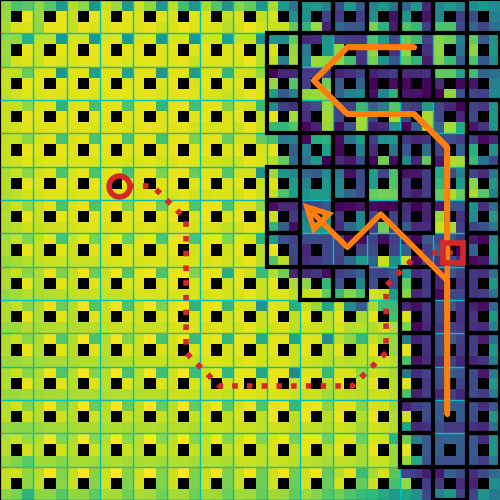}\hfill
  \includegraphics[width=0.33\columnwidth]{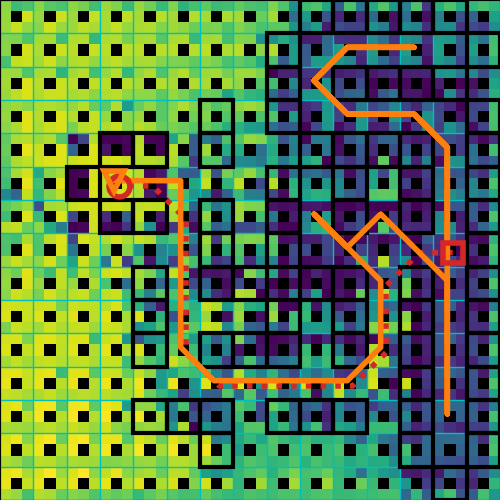}
  \includegraphics[width=0.33\columnwidth]{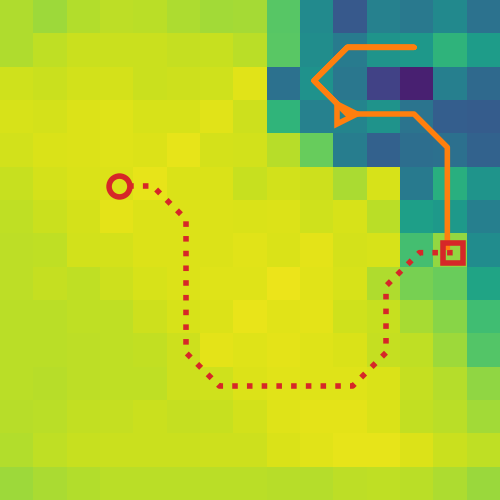}\hfill
  \includegraphics[width=0.33\columnwidth]{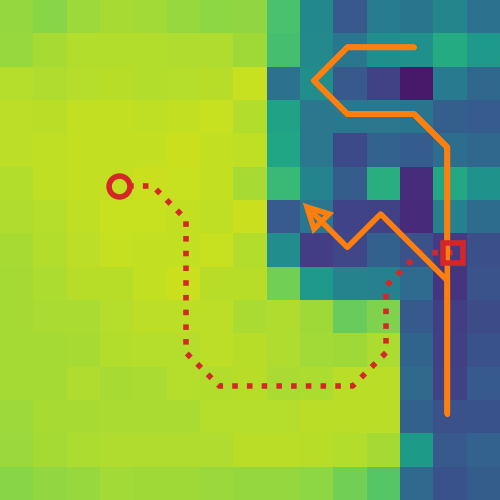}\hfill
  \includegraphics[width=0.33\columnwidth]{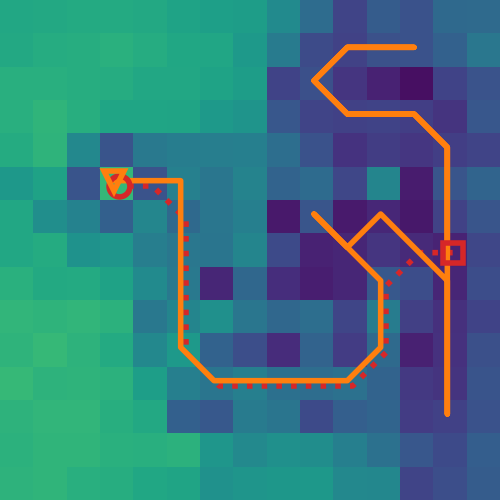}
  \caption{
    Example rollout of embodied CALVIN after $30$ steps (left column), $60$ steps (middle column) and $90$ steps (right column). CALVIN successfully terminated at $91$ steps.
    \textbf{(first row)} Input visualisation: unexplored cells are dark, the discovered target is yellow. The correct trajectory is dashed, the current one is solid. The orange triangle shows the position and the orientation of the agent.
    \textbf{(second row)} Predicted rewards (higher values are brighter). The 3D state-space (position/orientation) is shown, with rewards for the 8 orientations in a radial pattern within each cell (position). Explored cells have low rewards, while unexplored cells and the discovered target are assigned high rewards. 
    \textbf{(third row)} Predicted rewards averaged over the 8 orientations.
    \textbf{(fourth row)} Predicted values following the same convention. Values are higher facing the direction of unexplored cells and the target (if discovered). 
    \textbf{(fifth row)} Predicted values averaged over the 8 orientations.
  }
  \label{fig:calvin_emb_rollout}
\end{figure}

\subsection{Comparison of embodied navigation}
For visual comparison of CALVIN, VIN and GPPN, we generated a maze and performed rollouts using each of the algorithms, assuming partial observability and embodied navigation.

\begin{figure}
  \centering
  \includegraphics[width=0.33\columnwidth]{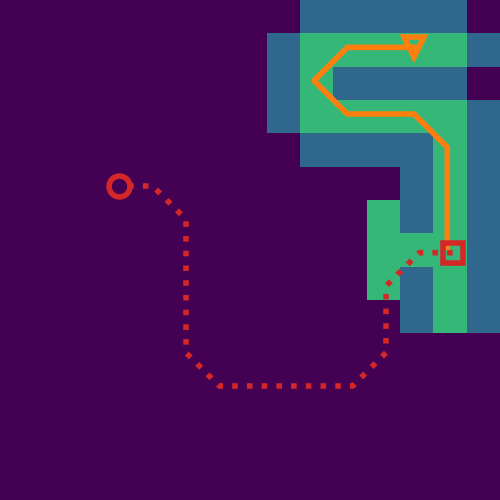}\hfill
  \includegraphics[width=0.33\columnwidth]{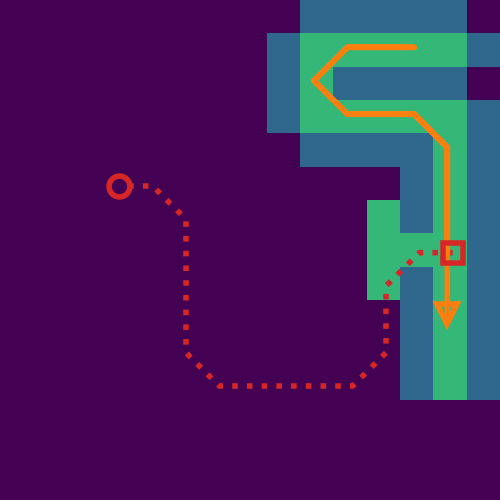}\hfill
  \includegraphics[width=0.33\columnwidth]{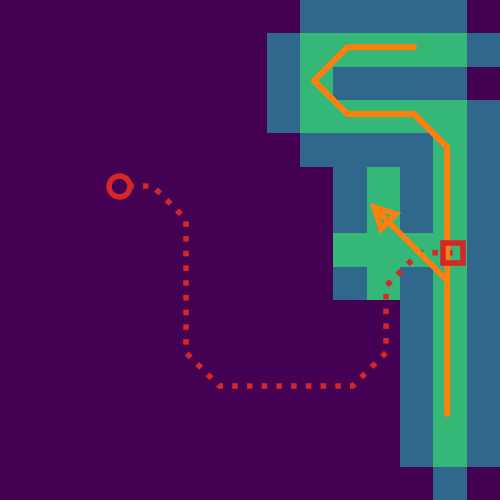}
  \includegraphics[width=0.33\columnwidth]{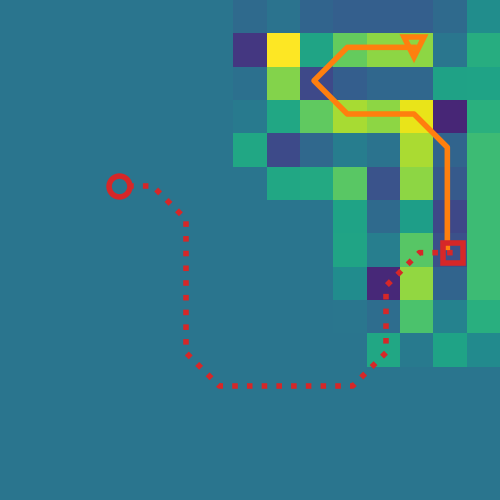}\hfill
  \includegraphics[width=0.33\columnwidth]{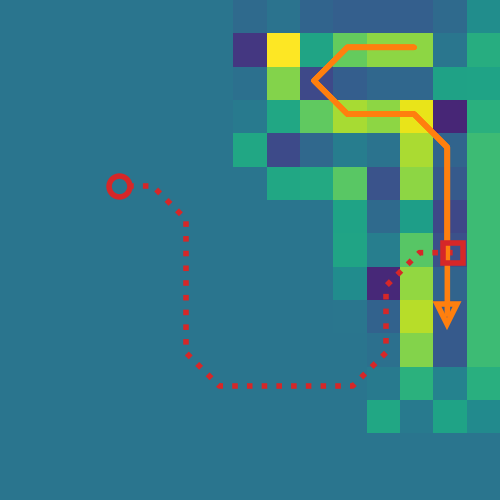}\hfill
  \includegraphics[width=0.33\columnwidth]{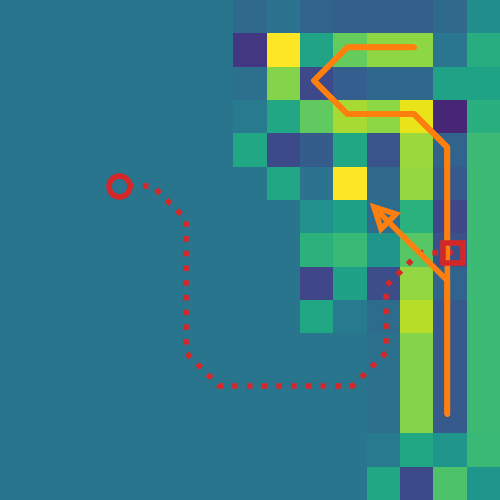}
  \includegraphics[width=0.33\columnwidth]{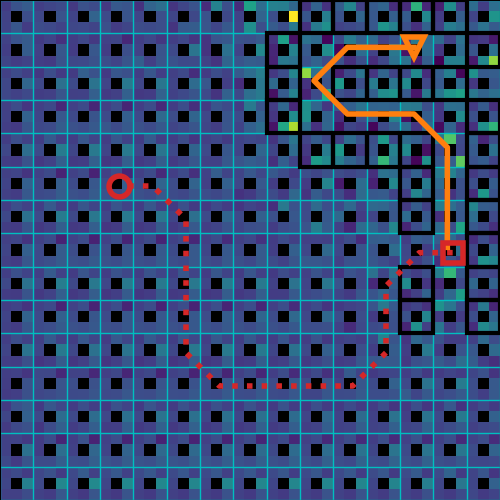}\hfill
  \includegraphics[width=0.33\columnwidth]{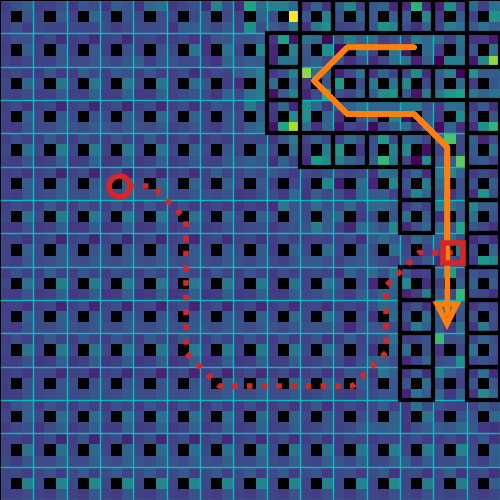}\hfill
  \includegraphics[width=0.33\columnwidth]{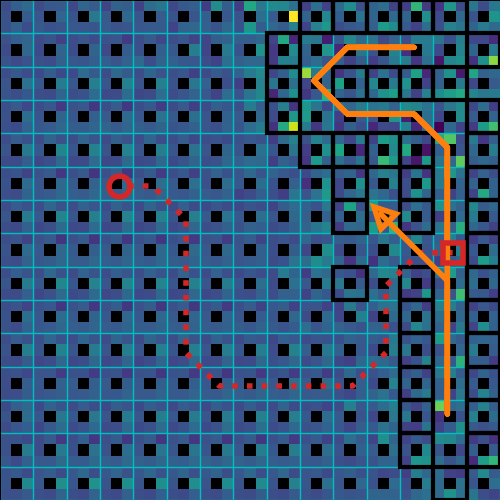}
  \includegraphics[width=0.33\columnwidth]{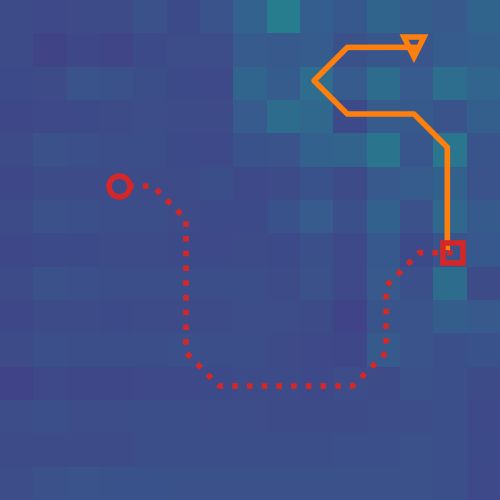}\hfill
  \includegraphics[width=0.33\columnwidth]{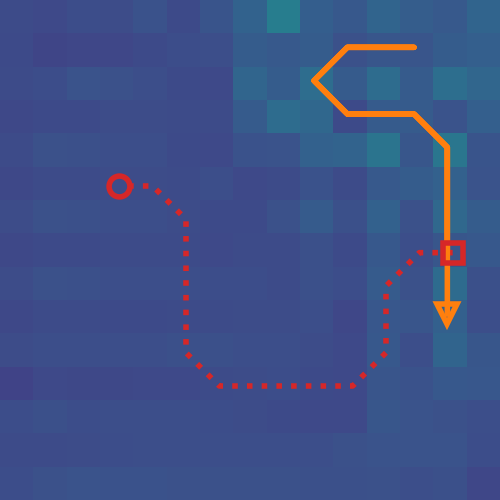}\hfill
  \includegraphics[width=0.33\columnwidth]{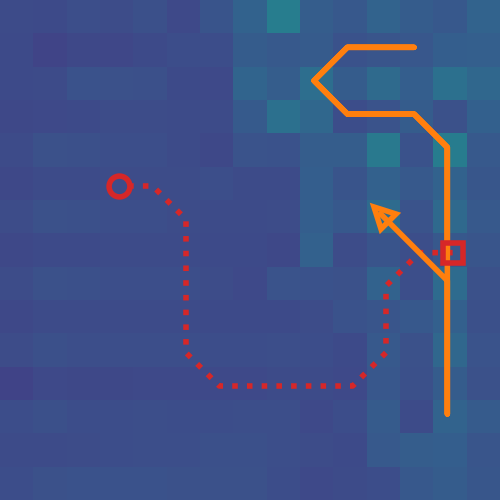}
  \caption{
    Example rollout of embodied VIN after $20$ steps (left column), $40$ steps (middle column) and $60$ steps (right column). VIN kept oscillating between the same two states after $57$ steps. The convention is the same as for \cref{fig:calvin_rollout}, except that a single reward map is shared across all orientations.
    \textbf{(first row)} Input visualisation.
    \textbf{(second row)} Predicted rewards.
    \textbf{(third row)} Predicted rewards averaged over the 8 orientations.
    \textbf{(fourth row)} Predicted values. 
  }
  \label{fig:vin_emb_rollout}
\end{figure}

\begin{figure}[t]
  \centering
  \includegraphics[width=0.33\columnwidth]{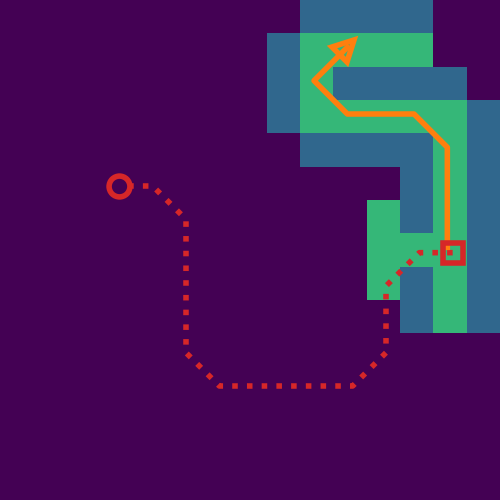}\hfill
  \includegraphics[width=0.33\columnwidth]{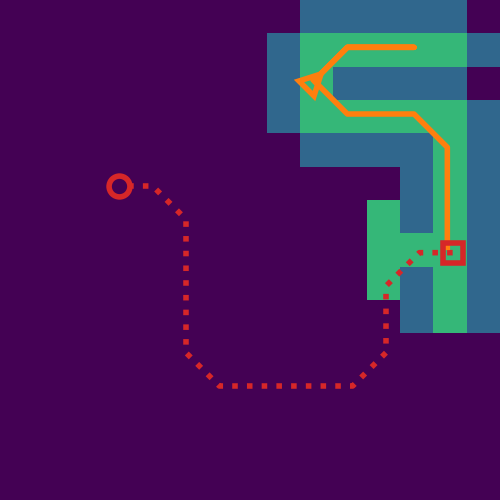}\hfill
  \includegraphics[width=0.33\columnwidth]{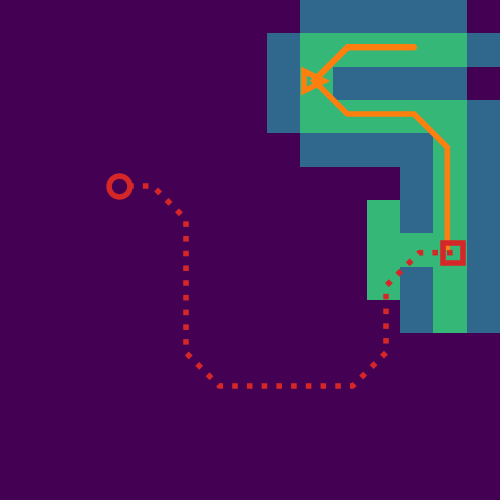}
  \includegraphics[width=0.33\columnwidth]{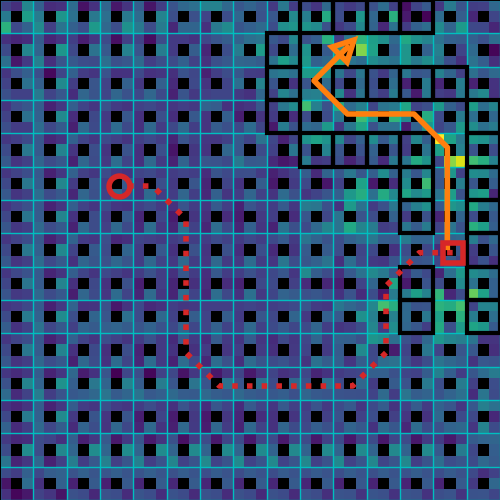}\hfill
  \includegraphics[width=0.33\columnwidth]{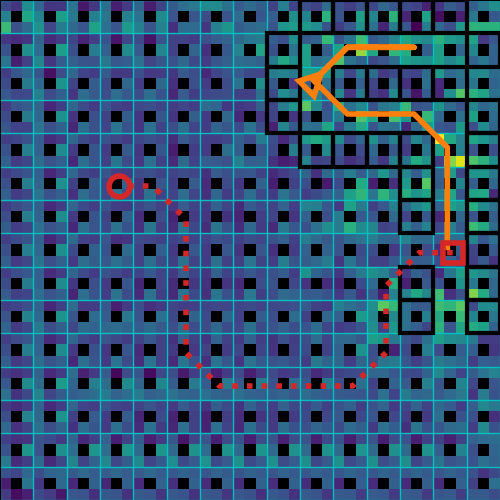}\hfill
  \includegraphics[width=0.33\columnwidth]{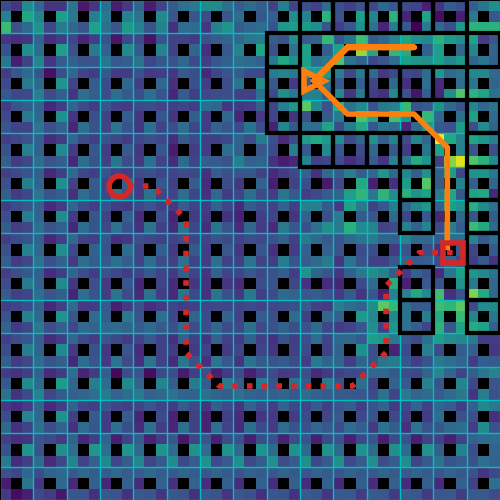}
  \includegraphics[width=0.33\columnwidth]{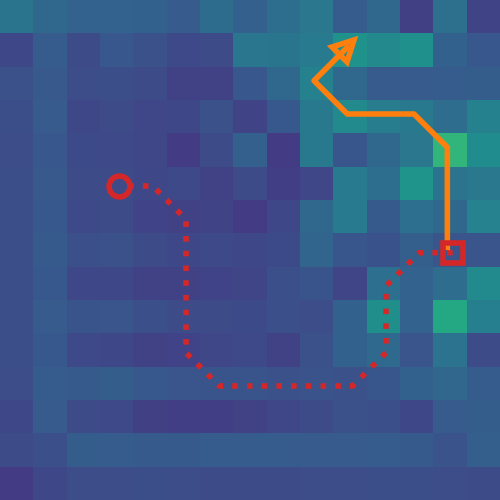}\hfill
  \includegraphics[width=0.33\columnwidth]{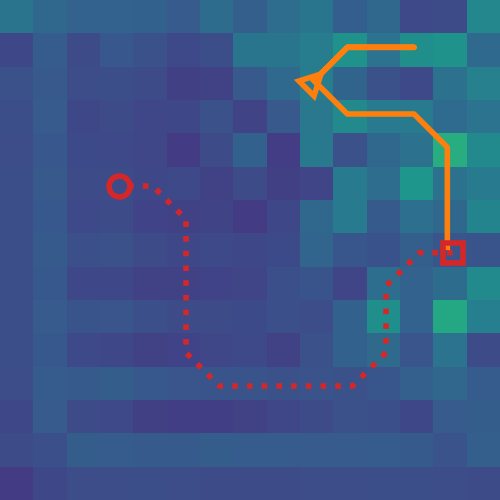}\hfill
  \includegraphics[width=0.33\columnwidth]{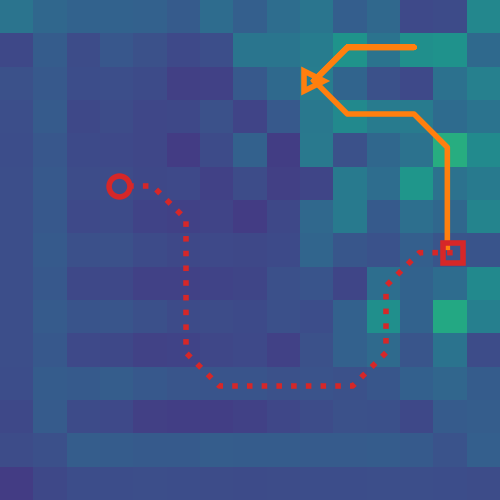}
  \caption{
    Example rollout of embodied GPPN after $15$ steps (left column), $30$ steps (middle column) and $45$ steps (right column). GPPN revisits the same sequences of states leading to a dead end after $45$ steps. The convention is the same as for \cref{fig:calvin_rollout}.
    \textbf{(first row)} Input visualisation.
    \textbf{(second row)} Predicted rewards.
    \textbf{(third row)} Predicted rewards averaged over the 8 orientations.
  }
  \label{fig:gppn_emb_rollout}
\end{figure}

\subsubsection{Rollout of CALVIN}

\Cref{fig:calvin_emb_rollout} shows an example of a trajectory taken by CALVIN at runtime, with corresponding observation maps, predicted values and predicted rewards for taking the ``done'' action. Similarly to \cref{sec:calvin_rollout}, the agent manages to explore unvisited cells and backtrack upon a dead end until the target is discovered.
One key difference is that now the agent learns to predict rewards and values for every discretised orientation as well as the discretised location. Upon closer inspection, we observe that the predicted values are higher facing the direction of unexplored cells and towards the discovered target. Since rotation is a relatively low cost operation, in this training example, the network seems to have learnt to assign high rewards to a particular orientation at unexplored cells, from which high values propagate. Rewards and values averaged over orientations yield a more intuitive visualisation.

\subsubsection{Rollout of VIN}
\label{sec:vin_rollout}

A corresponding visualisation for VIN is shown in \cref{fig:vin_emb_rollout}. Unlike CALVIN's implementation of rewards (eq. 5 in sec. 4.1) as a function of discretised states and actions, the ``reward map'' produced by the VIN does not offer a direct interpretation, as it is shared across all actions as implemented by Tamar \etal \cite{tamar_value_2016}, and is also shared across all orientations in the case of embodied navigation.
The values are also not well learnt, with some of the higher values appearing in obstacle cells. The unexplored cells are not assigned sufficiently high values to incentivise exploration by the agent. In this example, the agent gets stuck and starts oscillating between two orientations after 57 steps.

\subsubsection{Rollout of GPPN}

Finally, a visualisation for GPPN is shown in \cref{fig:gppn_emb_rollout}. Unlike VIN and CALVIN, GPPN does not have an explicit reward map predictor, but performs value propagation using an LSTM before outputting a final Q-value prediction. Similarly to \cref{sec:vin_rollout}, the values predicted is not very interpretable, and does not incentivise exploration or avoidance of dead ends. In this example, the agent keeps revisiting a dead end that has already been explored in the first 20 steps.

\end{document}